\pdfoutput=1
\documentclass[11pt]{article}

\usepackage[final]{acl}

\usepackage{times}
\usepackage{latexsym}

\usepackage[T1]{fontenc}
\usepackage[utf8]{inputenc}

\usepackage{microtype}

\usepackage{inconsolata}

\usepackage{graphicx}
\usepackage{booktabs}
\usepackage{todonotes}
\usepackage{latexsym}
\usepackage{subcaption}
\usepackage{xspace}
\usepackage{colortbl}
\usepackage{amssymb}
\usepackage{multirow}
\usepackage{dashrule}
\usepackage{makecell}
\usepackage{fix-cm}

\definecolor{cellground}{rgb}{0.82, 0.82, 0.82}

\usepackage{tcolorbox}
\usepackage{tabularx}

\usepackage{todonotes}

\makeatletter
\newcommand*\iftodonotes{\if@todonotes@disabled\expandafter\@secondoftwo\else\expandafter\@firstoftwo\fi}
\makeatother

\definecolor{ivanlime}{rgb}{0.9,1,0.3}
\definecolor{awesomered}{rgb}{1.0, 0.13, 0.32}

\newcommand{\rparagraph}[1]{\vspace{1.4mm}\noindent\textbf{#1.}}

\newcommand{\rrparagraph}[1]{\vspace{0.6mm}\noindent\textit{#1}}
\newcommand{\sparagraph}[1]{\vspace{0.0mm}\noindent\textbf{#1.}}

\definecolor{darkgreen}{rgb}{0.024, 0.631, 0.094}

\newcommand{\name}{ReCoVeR\xspace}
\newcommand{\nametrain}{ReCoVeR+\xspace}
\newcommand{\vonename}{ReCoVeR}
\newcommand{\mlpname}{ReCoVeR+}
\newcommand{\neuron}{Lang Neuron}
\newcommand{\monolc}{Mono-LC\xspace}
\newcommand{\crosslc}{Cross-LC\xspace}

\title{ReCoVeR the Target Language: \\ Language Steering without Sacrificing Task Performance}

\author{Hannah Sterz$^1$~~~~~Fabian David Schmidt$^2$~~~~~Goran Glavaš$^2$~~~~~Ivan Vulić$^{1}$ \\
$^1$University of Cambridge~~~~~$^2$University of Würzburg}

\begin{document}
\maketitle
\begin{abstract}
As they become increasingly multilingual, Large Language Models (LLMs) exhibit more \textit{language confusion}, i.e., they tend to generate answers in a language different from the language of the prompt or the answer language explicitly requested by the user. In this work, we propose \name (REducing language COnfusion in VEctor Representations), a novel lightweight approach for reducing language confusion based on language-specific \textit{steering vectors}. We first isolate language vectors with the help of multi-parallel corpus and then effectively leverage those vectors for effective LLM steering via fixed (i.e., unsupervised) as well as trainable steering functions. Our extensive evaluation, encompassing three benchmarks and 18 languages, shows that \name effectively mitigates language confusion in both monolingual and cross-lingual setups while at the same time---and in contrast to prior language steering methods---retaining task performance. Our data code is available at \url{https://github.com/hSterz/recover}. 
\end{abstract}

\section{Introduction}
% - introduce language confusion and why it is an important problem
%     - english centric
% - What are the short comings of existing methods
%     LSI
% - what do we do and why is it cool
%     easy to extend

%multilingual LLMs are cool
% Over the last few years, Large Language Models (LLMs) have \textit{de facto} become the default tool for addressing the vast majority of language-based tasks, owing to their impressive performance and generalization abilities in language understanding, reasoning, and language generation
% %, and .   translation, reasoning, and understanding tasks 
% \cite{achiam2023gpt, wei2022emergent, kojima2022large}. 

Large Language Models (LLMs) are becoming increasingly multilingual~\cite{aryabumi2024aya,team2025gemma}, progressively demonstrating more and more of their abilities across a broader set of natural languages. Broadening language support, however, increases the risk of \textit{language confusion} \cite{marchisio-etal-2024-understanding}: the phenomenon where an LLM answers in a language that is different from the language that the user---explicitly or implicitly---requested, or switches the language mid-reply. 
%%%
For instance, \newcite{marchisio-etal-2024-understanding} report GPT4o has a line-level pass rate of $88\%$ for the \crosslc Portuguese subset of the LCB. Thus, a non-negligible proportion of the generated lines are not in the expected language. While language confusion is not a frequent phenomenon in all languages, it has a detrimental effect on the user experience. Switching to a language that the user did not specify is likely to completely prevent the user from understanding the generated response. An inability to interact with users in their preferred language(s) hinders the global adoption of LLMs and---given that confusion is more prevalent for low-resource languages---exacerbates the exclusivity of state-of-the-art language technology. 
Such inconsistencies also complicate evaluation, as metrics that assume stable language output may overestimate true model performance.  
%and To ensure that LLMs are accessible for people from various backgrounds, LLMS must be able to interact with the user in their preferred language.
       
\begin{figure}[t]
    \centering
    \includegraphics[width=\linewidth]{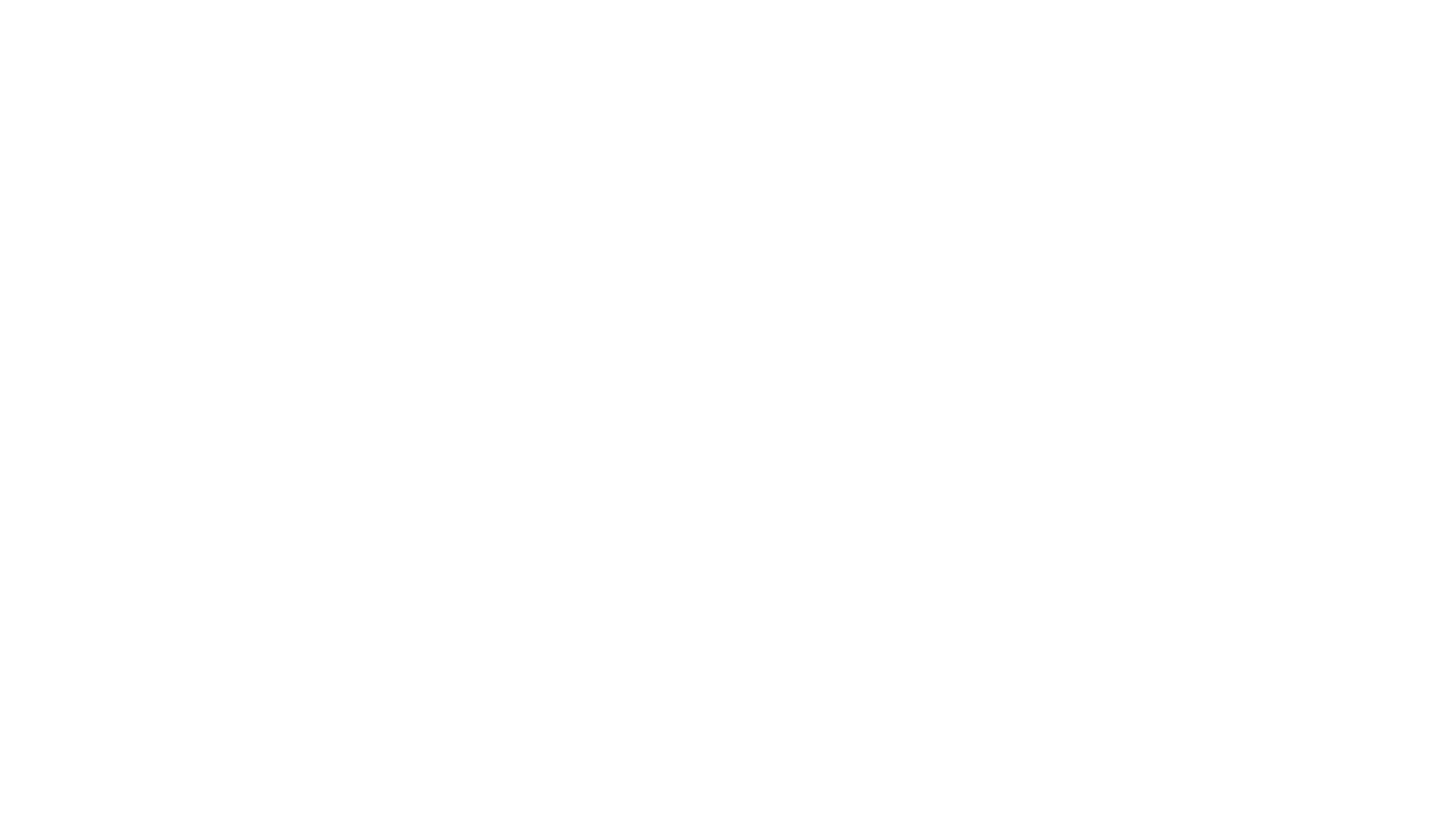}
    \caption{Generated answers for Llama 3.1 without and with ReCoVeR.}
    \label{fig:examples}
\end{figure}
%What is language confusion
%With multilingual LLMs, a phenomenon called language confusion (LC) has become more relevant. It describes when an LLM does not answer in the language the user (explicitly or implicitly) requested or if it switches to another language mid-reply \cite{marchisio-etal-2024-understanding}. 

%How do llms do out of the box (Why is it interesting to work on? monolingual vs multilingual

%What is other work LSI
Intuitively, language confusion can be somewhat mitigated with in-context examples (i.e., few-shot in-context learning, ICL) in the desired response language as well as via multilingual instruction-tuning \cite{marchisio-etal-2024-understanding}, but these assume existence of labeled data in each target language of interest. 
%%%
In contrast, \newcite{yunfan-etal-2025-mitigating} propose an unsupervised inference-time intervention to mitigate language confusion based on \textit{language vectors} that are added to hidden representations in the forward pass. 
%Why is LSI not enough?
While an inference-time solution, 
%for  that can be plugged in during inference,
this approach suffers from two key  drawbacks: (1) language vectors are computed jointly, i.e., dependent to one another, which means that all language vectors need to be recomputed from scratch when adding a new language; (2) while it improves language fidelity, it actually harms task performance (e.g., answer accuracy in question answering).  
%introducing new languages requires recomputing all language vectors from scratch. Moreover, LSI often harms performance on tasks, such as question answering or crosslingual summarisation, while reducing language confusion.

\begin{figure*}[t!]
    \centering
    \includegraphics[width=0.87\linewidth]{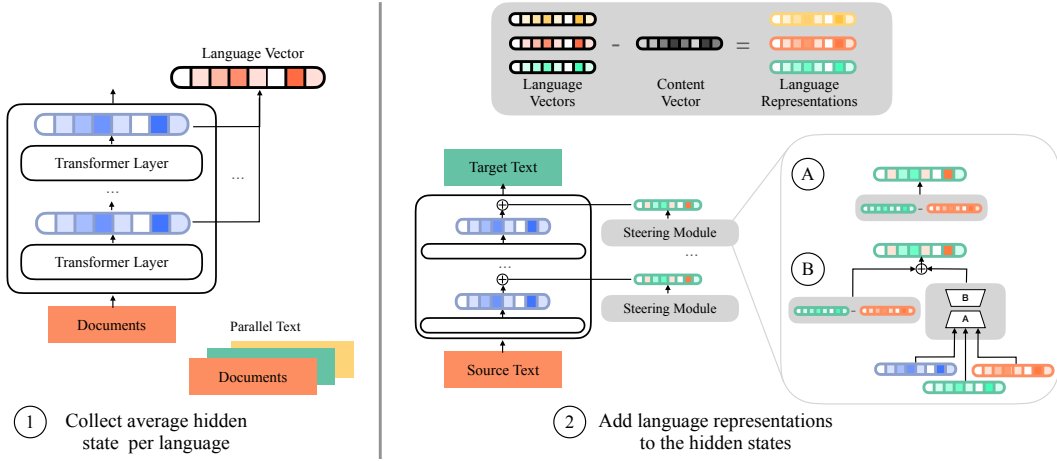}
    \caption{Illustration of \name. 1. The language vectors are computed based on the hidden representations on multi-parallel text. 2. The language representations are obtained by subtracting content from language vectors. These language representations can be used in A) unsupervised arithmetic steering and B) learned steering module. }
    \label{fig:illustrate-method}
    \vspace{-1.5mm}
\end{figure*}

%What do we do?
\rparagraph{Contributions} In this work, we introduce a novel lightweight approach for reducing language confusion with language vectors, dubbed \name (REducing language COnfusion in VEctor Representations), illustrated in Figure \ref{fig:illustrate-method}. We use multi-parallel data to pre-compute (1) language vectors for each Transformer layer (as average representations over all in-language input samples and all token positions) and then (2) language-agnostic, content vectors as averages of all language-specific vectors. By subtracting the content vectors from the language vectors, we obtain the language-specific representations (i.e., the \textit{steering vectors}), which finally allow us to mitigate language confusion at inference time by means of simple arithmetic operations. 
%in the style of $r_{target} - r_{source}$. 
%%%
We also experiment with trained interventions that learn how to compute the steering vector from the individual language vectors.
%\todo{What do we want to call our method? I like this acronym but not necessarily the full name}, 

Conceptually, like the concurrent work of \newcite{yunfan-etal-2025-mitigating}, \name adds (and subtracts) language-specific representations from intermediate token representations, but the way we compute and apply language vectors allows for seamless addition of new languages, without the need to re-compute language vectors for all existing languages (see Figure \ref{fig:examples} for example generations of Llama). Our extensive evaluation focuses on language confusion in both monolingual and cross-lingual setups and encompasses 18 typologically diverse languages. \name effectively reduces language confusion across the board, and crucially---unlike the concurrent approach \cite{yunfan-etal-2025-mitigating}---largely retains task performance.    
%comes with a major advantage of 
%which also relies on interventions that adapt hidden representations to reduce language confusion. But the way we compute and apply the language vectors differs from LSI. The setup is illustrated in Figure \ref{fig:illustrate-method}.  

%Wht are our results
%We evaluate language confusion and performance on downstream tasks. Language confusion alone is not sufficient, as the text can be in the correct language but not coherent. For instance, repeating a word in the target language would achieve perfect language confusion scores but prevent any meaningful answer. We will show that \name consistently improves crosslingual LC over zero-shot performance and LSI. 

%What are our contributions?
%Our contributions include: (1) proposing \name, a method to mitigate language confusion in LLMs, (2) exploring a module to learn steering vectors based on the language representations, and (3) a broad evaluation including monolingual and multilingual language confusion as well as the performance on downstream tasks. 

\section{Background and Related Work}

We first provide a brief overview of the body of work on manipulating LLMs via steering vectors and then describe the existing work on using steering vectors for mitigating language confusion.    

\rparagraph{Steering Language Models}
%%%
Hidden representations of LLMs conflate semantic content of the input text (i.e., prompt) with other aspects such as language, script, or style \cite{bricken2023towards}; the tokens that the LLM generates are predicted from representations that mix all these aspects. The body of work on \textit{representation engineering} aims to disentangle the contributions of different aspects in hidden representations, in order to accentuate or attenuate an aspect of interest, steering that way the behavior of an LLM in the desired direction \citep{zou2310representation, turner2023activation, wang2025semanticsadaptive}. 

Representation steering \cite{stolfo2024improving, stoehr-etal-2024-activation, zhang2025uncovering, subramani-etal-2022-extracting, hernandez2024inspecting}, as a predominant form of representation engineering, captures the representations of particular concepts (e.g., \textit{truthfulness} or \textit{toxicity}), as encoded in the representation spaces of LLM's layers, and then  modifies hidden representations of the input with these concept representations: this has the goal of steering the model behavior in the direction of the concept (e.g., towards more truthful or less toxic generations). 
%%%
The success of representation steering can be explained by the linear representation hypothesis, which posits hidden representations to be linear combinations of aspect/concept vectors \cite{park2024linear}. The most common ways of computing concept vectors is (i) by contrasting representations obtained from a set of positive examples which exhibit the desired behavior (e.g., non-toxic generations) against those obtained from a set of negative examples which do not exhibit the desired behavior (e.g., toxic generations) 
%of the desired behavior (e.g., toxic vs. non-toxic generations) and using the difference between the representations of the samples as the concept vector 
\cite{jorgensen2023improving, rimsky-etal-2024-steering, cao2024personalized}
or (ii) by means of linear probing, identifying the most concept-sensitive dimensions of the hidden representations \cite{alain2017understanding,yunfan-etal-2025-mitigating}.
The desired behavior is then induced by adding (or subtracted) the concept vectors from from hidden states at inference time \citep{singh2024representation, liu2023context}.
%For example, adding the truthfulness vector increases the truthfulness of the generated output.

%Steering can help not only to improve fidelity. \cite{wang2024bridging} learn a projection matrix $W^*$ that aligns the hidden states of language pairs. Applying the projection during inference enables crosslingual generation. However, inference requires the matrix $W^*$ for every language pair, resulting in $|L|^2$ matrices. Moreover, the hyperparameters are language pair , requiring $|L|^2$ hyperparameter searches.

\rparagraph{Steering to Mitigate Language Confusion}
The expected language of an LLM's response is given either (i) implicitly, where the model is expected to provide the answer in the language of the prompt or (ii) explicitly, with the language of the response specified as part of the prompt (e.g., \emph{'What is the capital of France? Answer in German!'}), i.e., the expected answer language differs from the prompt language. Throughout the paper, we refer to the former as monolingual language confusion (\monolc) and denote the latter as cross-lingual language confusion (\crosslc). 
%if the prompwith the answer la language can be indicated in two ways. First, the language can be implicitly specified through the prompt language. The model should respond in the language used in the prompt if there is no explicit instruction (termed \textbf{monolingual} LC). Second, a different language can be specified in the prompt, e.g.  (termed \textbf{crosslingual} LC).
Intuitively, state-of-the-art LLMs suffer from language confusion more in cross-lingual than in monolingual setups~\cite{marchisio-etal-2024-understanding}. % \missing{[reference]}. 
%perform better in the monolingual setting than in the crosslingual setting. 
As we show in \S\ref{sec:results}, owing to their English-centric nature, LLMs especially struggle with \crosslc when the prompt is not in English.
%crosslingual sample is not in English, probably due to the English-centric nature of LLMs.

Language confusion occurs when the model answers in the wrong language or switches between languages during generation \cite{marchisio-etal-2024-understanding}. \newcite{tang-etal-2024-language} find that specific neurons encode the language information, which points to representation steering as a suitable framework for mitigating language confusion: if we can find reliable concept vectors for languages, we could steer the generations towards the target language. 
%to reduce language confusion. 
%%
Following this intuition, in their language steering at inference (LSI) approach  \newcite{yunfan-etal-2025-mitigating} capture language vectors by identifying the most language-sensitive dimensions in the hidden states via linear probing. For each language $l$, LSI obtains a \textit{representation mask} $M_l$ that describes which dimensions are relevant for correctly predicting $l$. Next, LSI determines the difference between hidden representations when prompted with an instruction-only target language prompt $V_l^*$ and a prompt with an additional in-context example $V_l$, following \cite{marchisio-etal-2024-understanding}, who report that in-language examples in the context reduce language confusion. $V_l$ is the language mask applied to the hidden state: $V_l = M_l \odot h $.
%%%
They then compute the language representation (layer $i$):
\begin{equation}
 r_l^{(i)} = \frac{1}{K} \sum_{k=1}^N V_{l,k}^{(i)}, - V_{l,k}^{(i)*}.
 \end{equation}
Here, $K$ is the number of prompts. Intuitively, this formula reduces the tendency to answer in the dominant language, i.e., English.
Finally, to steer the representations at inference towards the target language, they add $r^{(i)}_l$ (scaled with the hyperparameter $\gamma$) to the corresponding hidden representations $h^{(i)}$: $\hat{h}^{(i)} = h^{(i)} + \gamma r_l^{(i)}$. 

% 1 need to train new probe if new language is added
% 2 
LSI comes with three prominent drawbacks. First, being based on the language classification probe to produce language-specific masks, LSI produces language vectors $r_l$ that directly depend on the set of languages for which the probe is trained; this prevents an easy post-hoc addition of new target languages as that requires retraining the probe and recomputing $r_l$ vectors for all existing languages too. Second, language representations $r_l$ are computed focusing on the dominant LLM language: this, however, limits the applicability of LSI as it cannot be used for mitigating \crosslc for non-English prompts. Finally, the reduced language confusion that LSI achieves comes at the cost of a substantial loss of task performance (see \S\ref{sec:results}). We believe that this is because LSI only steers (i.e., changes) hidden representations in a subset of dimensions (i.e., the most language-specific dimensions, as identified by the probe). This makes the steered representations more likely to fall out the output representation distribution of the layer and as such they ``confuse'' the rest of the model. 
%which makes  task, it also  significant drops in task-performance   changing only some of the dimensions of representations vector the most prominent language-specific dimensions .

%\subsection{Representation Engineering}

\section{Lightweight Language Steering}

We introduce \name (REducing language COnfusion in VEctor Representations), a lightweight language steering approach that addresses the shortcomings of the existing approaches. \name uses a (readily available) multi-parallel dataset with a large language coverage to estimate language-specific vectors and, unlike ICL- and fine-tuning-based mitigation of LC~\cite{marchisio-etal-2024-understanding}, does not require labeled task data in target languages. \name computes language representations in relation to the average representations across all languages: as such, \name is able to support \crosslc with non-English prompts. Importantly, \name steering vectors for new languages can easily be computed post-hoc, without the need to recompute the steering vectors of existing languages, which is one of the key drawbacks of LSI \cite{yunfan-etal-2025-mitigating}.       

%We propose \name, a steering method designed to reduce language confusion while maintaining (or improving) performance.

Steering comprises two sub-problems. The first one is isolating the vector that captures how a concept (in our case, a language) is encoded in LLM's hidden representations (\S\ref{ssec:find_vectors}). 
%%%
%We describe our approach in \S\ref{sec:find_vectors}.
The second subproblem is designing the steering strategy, i.e., how to use the concept (language) vectors to manipulate the hidden representations so that the model generates text with the desired properties (in our case, in the desired target language). To this end, we explore both unsupervised (\S\ref {ssec:apply_vectors}) and supervised strategies (i.e., learning how to combine language vectors for optimal steering; \S\ref {ssec:learned_vectors}). 

\subsection{Isolating Language Vectors}
\label{ssec:find_vectors}

We start from the following intuitive hypothesis: differences in hidden representations obtained for a pair of parallel texts, i.e., texts that are mutual translations, primarily stem from how the LLM encodes the input language.
Because of this, we propose to use a multi-parallel corpus to isolate language representations. 
For each language $l \in L$ covered by the multi-parallel corpus $D_L$, we compute one language vector $v^{(i)}_l$ for each transformer layer $i$ (from 1 to $N$) as the mean of the hidden representations of layer $i$, averaged across all tokens of all input samples in language $l$ (i.e., across all tokens of $D_l$, the monolingual portion of $D_L$): 

$$ v^{(i)}_l =\frac{1}{|D_l|} \sum_{x\in D_l}  \frac{1}{P}\sum_{p=1}^P h^{(i)}_p(x) $$

where $h^{(i)}_p$ denotes the hidden representation of the $p$-th token of the input prompt $x$ at the output of the $i$-th layer.\footnote{We exclude the representations of the first token in each sample, i.e., the \emph{'beginning of sequence'} token, assuming that its representation encodes the input language less prominently.} 
The obtained language vectors $v^{(i)}_l$, however, still conflate the representation of the language $l$ with the aggregate representation of the ``content'' of our multi-parallel corpus $D_L$. In order to obtain language representations $r^{(i)}_l$ that are to be effectively used for language steering, we first need to eliminate the corresponding ``content'' representations $c^{(i)}$ from respective language vectors $v^{(i)}_l$. 
To this end, we assume that we will obtain the language-agnostic content vector $c^{(i)}$ by averaging language vectors $v^{(i)}_l$ across all languages $l \in L$:

$$ c^{(i)} = \frac{1}{|L|}\sum_{l \in L} v^{(i)}_l $$
We then obtain the language representations by subtracting the content vectors from respective language vectors: $r^{(i)}_l = v^{(i)}_l - c^{(i)}$.
We next seek to exploit the language representations $r_l$ computed from the multi-parallel corpus $D_L$ to steer the LLM towards generations in the target language $l$.

\subsection{Unsupervised Language Steering}
\label{ssec:apply_vectors}
 
%The next question is how can we use them to reduce the language confusion of an LLM. 
We are looking to integrate $r^{(i)}_l$ into the hidden representations, output of the $i$-th layer, in a manner that fulfills two mutually conflicting objectives: (1)   
we need \textit{enough} information from $r^{(i)}_l$ to steer the generation towards $l$ and at the same time (2) change the hidden representations \textit{as little as possible}, in order to 
%At the same time, the manipulation needs to be small enough to 
prevent model collapse. We find that, for \monolc, L2-normalizing the $r_l$ and scaling it (with a hyperparameter $\alpha$) yields steering vectors that can achieve both goals: 
\begin{equation} 
\label{eq:mono}
\hat{h}^{(i)} = h^{(i)} + \alpha \frac{r^{(i)}_{l}}{|r^{(i)}_{l}|}
\end{equation} 

For \crosslc, where the prompt is in a \textit{source} language and the expected answer is in a different \textit{target} language, we need an additional steering component to discourage the LLMs from generating the answer in the language of the prompt. To this end, we set our final steering vector to the difference between $r^{(i)}_{target}$ and $r^{(i)}_{source}$, i.e., we steer the representations towards the target language \textit{and} away from the source language (and we again L2-normalize and scale the steering vector): 
%one for the source and one for the target language. To prevent the model from generating text in the source language, we perform language arithmetic. That is, we subtract the source representation from the target representation.
%With this, we get the following steering function:
%
\begin{equation} \label{eq:cross} 
 \hat{h}^{(i)} = h^{(i)} + \alpha \frac{r^{(i)}_{target} - r^{(i)}_{source}}{|r^{(i)}_{target} - r^{(i)}_{source}|}
 \end{equation}
\noindent Note that, due to the fact that \name does not compute language representations $r_l$ relative to any fixed dominant/pivot language, our steering via above language arithmetic applies to arbitrary pairs of source and target languages in \crosslc. 
%transfers to new language pairs without further data.

We modify hidden representations $h^{(i)}$ according to Eq.\,\ref{eq:mono} (\monolc) and Eq.\,\ref{eq:cross} (\crosslc) for all tokens in the input sequence except the first token, following our assumption that representations of the sequence start token do not encode language components.
%not contain language-specific information, we exclude it from this operation. The hidden state of the first token remains unchanged.
We additionally hypothesize that retaining the norm of hidden representations after steering may be important for preventing model collapse. We thus introduce an additional binary-value hyperparameter that decides whether to restore the representation norms after steering, i.e., whether to ensure that $||\hat{h}^{(i)}|| = ||h^{(i)}||$.

\subsection{Learning Language Steering}
\label{ssec:learned_vectors}

We next investigate if we can learn, in a sample-efficient manner, a steering function that is more effective than our simple unsupervised steering from \S\ref{ssec:apply_vectors}. 
Aiming for sample-efficient training, we choose our steering function to be a low-rank intervention (with a residual), to which we input our language representation(s) $r^{(i)}_l$ concatenated to a hidden representation $h^{(i)}$: 

$$ h_{+}^{(i)} =\hat{h}^{(i)} + AB[h^{(i)};r^{(i)}_{target}; r^{(i)}_{source}]$$

\noindent with $A \in \mathbb{R}^{3d\times r}, B \in \mathbb{R}^{r\times d}$ as trainable parameters of the steering function. For training stability, as common in low-rank interventions, we zero-initialize the up-projection $B$. $\hat{h}^{(i)}$ is defined as:
\begin{equation} 
 \hat{h}^{(i)} = h^{(i)} + \alpha \frac{r^{(i)}_{target} - \beta r^{(i)}_{source}}{|r^{(i)}_{target} - \beta r^{(i)}_{source}|}
\end{equation}

The above formula applies to \crosslc and \monolc . To make it compatible with \monolc (where $r^{(i)}_{target} = r^{(i)}_{source}$) we introduce parameter $\beta$ to scale the source representation:

\section{Experimental Setup}
\label{sec:expsetup}

\sparagraph{Obtaining Steering Vectors} To isolate the language representations $r_l$ (\S\ref{ssec:find_vectors}), we need a multi-parallel corpus $D_L$: here, we resort to FLORES-200 \cite{costa2022no}, a sentence-level multi-parallel dataset covering 200 languages. 

We additionally require a training dataset for our learnable steering function (\S\ref{ssec:learned_vectors}). This dataset should (1) contain both monolingual as well as cross-lingual prompt-response pairs and (2) cover a wide range of typologically diverse languages (and a wide range of tasks), in order to enable cross-lingual generalization of our learned steering function. We obtain such a dataset by first sampling 4400 single-turn instances (i.e., prompt-response pairs) from the instruction-tuning dataset Tulu v3 \cite{lambert2024tulu} and then translate each of them (with GPT-4o) by sampling the translation language for the prompt and (independently) for the response from the following set of languages: English, Spanish, French, German, Portuguese, Russian, Chinese, Japanese, Arabic, Hindi, Indonesian, Hebrew, Tamil, Farsi, Thai, Polish, Dutch, Bengali.   %%   
%well allow to both perform cross-lingual and monolingual generation. This dataset should cover a wide range of languages to enable generalization. As this should cover a wide range of questions and tasks, we choose instruction following as the task. Thus, Tulu v3 \cite{lambert2024tulu} is well-suited as a base for this dataset.
%%% 
%We sample 4400 instances that consist of only one turn. Using GPT-4o, we translate these samples to source and target language pairs sampled from:
%English, Spanish, French, German, Portuguese, Russian, Chinese, Japanese, Arabic, Hindi, Indonesian, Hebrew, Tamil, Farsi, Thai, Polish, Dutch, Bengali. These pairs also include monolingual pairs with identical source and target languages. 
We increase the likelihood of sampling monolingual pairs to ensure that they are sufficiently represented.
We deliberately exclude from translation four other languages present in the evaluation benchmarks (Italian, Korean, Turkish, and Vietnamese) in order to test our learned steering for (zero-shot) cross-lingual generalization.\footnote{Our final multilingual/cross-lingual instruction tuning dataset contains 4400 instances covering 18 languages: we provide the number of samples per language in Table \ref{tab:dataset_stats} and the training details in \S\ref{sec:trdetails_recoverplus}.} For details on the dataset and translation quality see \ref{app:dataset}.

\rparagraph{Evaluation Benchmarks}
We employ three different benchmarks for measuring language confusion. Two of them come with task-specific annotations, allowing us to also measure the impact of our language steering on task performance. 
%
%In order to, evaluate language confusion and performance of the LLMs we evaluation two language confusion benchmarks, where the second one as a question answering benchmark also includes the accuracy of correctly answered questions. Moreover, we add a crosslingual summarisation dataset to our evaluation to determine the impact of baselines and \name.

\rrparagraph{Language Confusion Benchmark} (LCB) \cite{marchisio-etal-2024-understanding} covers 15 languages and serves to measure language confusion.
%i.e., the (in)ability to consistently answer in the correct language. 
The corresponding metrics are Line-Level Pass Rate (LPR) and Word-Level Pass Rate (WPR). For LPR, the LLM generations are split into lines 
and each line is classified by the language identification classifier; LPR is then simply the percentage of lines in the requested/expected language. WPR is the proportion of words in the correct lines (identified for LPR) in the correct script\footnote{Determining the language from a single word is challenging, as the same words can occur in multiple languages. The script of the word serves as an approximation. As a result, WPR is only applicable to languages with non-Latin scripts.}.
%WLC is the percentage of words in the correct language. Here,the correct language is approximated with the correct script. The model tends to fall back to English; therefore, the approximation works for non-Latin scripts. Thus, WLC is only available for languages with a non-Latin script.
LCB consists of two portions: monolingual (LLM expected to reply in the language of the prompt) and cross-lingual (LLM instructed to reply in the specified language, different from the prompt language). 
%The benchmark consists of two subsets: the monolingual set contains prompts in the target language and expects the model to answer in the same language. The cross-lingual set contains English prompts and specifies the target language, e.g. 'Answer in German!'. 

\rrparagraph{MultiQ} \cite{holtermann-etal-2024-evaluating} is a multi-parallel QA dataset covering 137 languages. We use it to evaluate language confusion and QA accuracy\footnote{We only consider correct answers in the expected language as correct.}. For the latter, as in the original work, we evaluate with LLM-as-a-judge (GPT4o). We select 5 MultiQ languages for the \monolc evaluation. We also create cross-lingual instances with non-English prompts by using prompts in German, Basque, Farsi, French, Swahili, Turkish, and Chinese, including the translation of the cross-lingual instruction \textit{'Answer in X!'}, with X as the target language.

\rrparagraph{CrossSum} \cite{bhattacharjee-etal-2023-crosssum} is a cross-lingual summarization benchmark. encompassing more than 1500 language pairs. 
%Crosslingual summarization is the task of summarizing a given text in a source language in a different target language. CrossSum contains more than 1500 language pairs. 
Due to computational constraints, we limit our evaluation to four language pairs: English-Spanish, Spanish-French, French-Turkish, and Turkish-Swahili.

\rparagraph{Models}
We experiment with three different open multilingual instruction-tuned LLMs of varying size and declared language support: Llama 3.1 \cite{grattafiori2024llama} (8B model; officially supports 7 languages), Qwen 2.5 \cite{yang2024qwen2} (7B model; 29 languages), and Gemma 2 \cite{team2024gemma} (2B; no.~languages undeclared). 

Besides against the original model (i.e., without any language steering), we compare the performance of \name against the steering with LSI \cite{yunfan-etal-2025-mitigating} and language-specific neurons\cite{kojima-etal-2024-multilingual}. For LSI, we obtain the language masks $M_l$ for with samples from WikiLingua \cite{ladhak-etal-2020-wikilingua} and for languages not covered by WikiLingua, we use texts from Wikipedia\footnote{Obtained from \url{https://huggingface.co/datasets/wikimedia/wikipedia}}. We provide further details on LSI parameters in Table \ref{tab:hyperparams_lsi}. We evaluate our unsupervised steering (\S\ref{ssec:apply_vectors}; denoted as \name) as well as our learned steering functions (\S\ref{ssec:learned_vectors}, denoted as \nametrain). %The selected models vary in size and how many languages they officially support. This gives us detailed insight into the impact of our method.

%\textbf{Llama 3.1} \cite{grattafiori2024llama}: The Llama model is an open model that officially supports  German, French, Italian, Portuguese, Hindi, Spanish, and Thai. We use the \texttt{meta-llama/Llama-3.1-8B-Instruct} checkpoint of the model.

%\textbf{Qwen 2.5} \cite{yang2024qwen2}: Qwen is similar in size to Llama and claims to support over 29 languages. This makes it the most multilingual out of the three. We use the \texttt{Qwen/Qwen2.5-7B-Instruct} checkpoint.

%\textbf{Gemma 2} \cite{team2024gemma} Gemma is another open LLM. It does not explicitly state which languages other than English it supports, but it was designed with multilingual use-cases in mind. We use the smaller \texttt{google/gemma-2-2b-it} checkpoint.

%We compare \name with the zero-shot performance of the models and LSI \cite{yunfan-etal-2025-mitigating}. We obtain the language masks $M$ for LSI with samples from WikiLingua \cite{ladhak-etal-2020-wikilingua} and for languages not covered by WikiLingua, we use texts from Wikipedia \footnote{Wikipedia texts are obtained from \url{https://huggingface.co/datasets/wikimedia/wikipedia}}. The detailed hyperparameters used in the intervention can be found in Table \ref{tab:hyperparams_lsi}

\section{Results and Discussion}
\label{sec:results}

We first show and discuss the language confusion and task performance in \monolc and \crosslc setups and then provide further analyses for some of the design dimensions for language steering.  
%In the following, we will discuss the performance of baselines and \name in various scenarios. We look at the standard monolingual (\S\ref{sec:monolingual_results}) and crosslingual (\S\ref{sec:crosslingual_results}) setups, also including performance on question answering and crosslingual summarization tasks. 
%%Then, we investigate the impact of the explicit language information given in the crosslingual prompts (\S\ref{sec:crosslingual_no_lang_results}). 

\subsection{Language Confusion}
%\label{sec:monolingual_results}

\begin{table}[t]
\centering
\scriptsize
\def\arraystretch{0.85}
\setlength{\tabcolsep}{17pt}
%\resizebox{\linewidth}{!}{
\begin{tabular}{lrrr}\toprule
 & LCB &  \multicolumn{2}{c}{MultiQ}  \\ \cmidrule(lr){2-2} \cmidrule(lr){3-4} 
Model & LPR  & LPR & Acc \\
\midrule \rowcolor{gray!20} 
LLama 3.1 & 98.7 & 94.5 & \textbf{64.4} \\ 
%\midrule
+ LSI & 99.0 & 95.7 & 52.8 \\ 
+ \neuron & \textbf{99.2} & - & - \\
+ \vonename & 99.1 & 93.8 & 61.3  \\
% + \vonename &  \textbf{99.0} & 93.8 & 62.6 \\
% + v2 &  \\
+ \mlpname & 99.1 & \textbf{95.8} & 62.1 \\ % & 98.9 & 94.9 & 48.7 \\ 
\midrule \rowcolor{gray!20} 
Qwen 2.5 &  98.3 & 90.7 & 61.8\\
%\midrule
+ LSI & 98.0 & 92.7 & 51.0\\
+ \neuron & 96.2 & - & -  \\
+ \vonename & 97.7 & 92.0 & 62.7 \\
% + \vonename & 98.2 & \textbf{99.2} & \textbf{62.1} \\
% + v2 & \\
+ \mlpname & \textbf{98.5} & \textbf{93.2} & \textbf{66.5}\\ 
% 95.6 & 92.1 & 61.9\\
\midrule \rowcolor{gray!20} 
Gemma 2 & 88.4 & 91.8 & 38.5
 \\
+ LSI & 90.2 & \textbf{93.5} & 34.8 \\
+ \neuron & 89.7 & - & -  \\
+ \vonename & 87.8 & 91.6 & 38.3 \\
% + \vonename & 87.7 & 92.2 & 38.2\\
% + v2 & & \\
+ \mlpname & \textbf{98.1} & 92.9 & \textbf{47.4}\\ 
%\textbf{97.6} & \textbf{94.0} & \textbf{44.6} \\
\bottomrule
\end{tabular}
%}
\vspace{-1mm}
\caption{\monolc results on LCB and MultiQ.}
\label{tab:mono_agg}
\vspace{-2mm}
\end{table}

\begin{table}[th!]
    \centering
    \small
    \begin{tabular}{lccccc}
        \toprule
         & de & es & id & sw & zh\\
         \midrule
         \rowcolor{gray!20} Llama 3.1 & 52.8 & 55.3 & 50.2 & 38.3 & 51.3 \\
         + \name & 52.8 & 54.9 & 50.2 & 38.6 & 51.5 \\
         \bottomrule
    \end{tabular}
    \caption{MMLU accuracy for Llama 3.1 out-of-the-box and with \name.}
    \label{tab:mmlu}
\end{table}
\begin{table*}[t]\centering
\centering
\def\arraystretch{0.83}
\fontsize{8.2pt}{8.1pt}\selectfont
\resizebox{\linewidth}{!}{
%\scriptsize{
\begin{tabular}{clrrrrrrrrrrrrrrrr}\toprule
& Model  & ar	& de & es & fr & hi & id & it & ja & ko & pt & ru & tr & vi & zh & avg \\
\midrule \rowcolor{gray!10}
\cellcolor{white}& LLama 3.1 & 90.4 & 95.2 & 95.6 & 94.0 & 91.9 & 89.3 & 93.9 & 78.2 & 90.3 & 91.9 & 90.2 & 95.3 & 94.6 & 83.2 & 91.0  \\
& + LSI & 92.8 & 96.9 & 94.9 & 96.0 & 96.5 & \textbf{90.9} & 97.0 & 82.4 & 92.0 & 96.7 & 92.6 & 92.8 & 94.8 & 84.2 & 92.9 \\
& + \neuron & 94.3 & 95.2 & 95.0 & 95.0 & 95.3 & 86.3 & 95.6 & 89.5 & 92.1 & 92.2 & 95.6 & 97.3 & 98.0 & 89.6 & 93.6 \\
& + \vonename & 99.2 & \textbf{99.7} & \textbf{98.6} & \textbf{99.7} & 99.3 & 90.1 & \textbf{98.6} & \textbf{97.6} & \textbf{100.0} & \textbf{97.0} & 99.7 & \textbf{97.9} & \textbf{99.0} &\textbf{94.8} & \textbf{97.9} \\
% & + \vonename & 95.4 & 97.2 & 95.5 & 95.6 & 96.3 & 85.1 & 97.3 & 95.6 & 96.8 & 92.6 & 98.6 & 96.2 & 92.2 & 88.9 & \textbf{94.5}\\
% & + v2 & 80.0 & 97.7 & 97.3 & 95.6 & 90.0 & 82.5 & 97.3 & 90.2 & 88.4 & 93.6 & 91.2 & 94.6 & 94.6 & 87.6 & 91.5 \\
& +\mlpname & \textbf{99.6} & 98.3 & 98.3 & 98.7 & \textbf{100.0} & 88.2 & \cellcolor{cellground}96.9 & 97.3 & \cellcolor{cellground}99.7 & 94.2 & \textbf{100.0} & \cellcolor{cellground}96.6 & \cellcolor{cellground}97.3 & 93.6 & 97.0\\ 
% 100.0 & 99.3 & 99.0 & 97.9 & 99.7 & 88.7 & 98.6 & 97.3 & 99.7 & 93.9 & 99.7 & 99.7 & 99.3 & 91.1 & 97.4\\%99.6 & 95.9 & 97.9 & 95.5 & 99.0 & 90.7 & \cellcolor{cellground}57.4 & 97.3 & \cellcolor{cellground}94.6 & 94.4 & 99.3 & \cellcolor{cellground}54.4 & \cellcolor{cellground}84.3 & 92.1 & 89.5  \\
\cmidrule{2-17}  \rowcolor{gray!10}
\cellcolor{white} & Qwen 2.5 & 93.2 & 95.5 & 94.7 & 93.4 & 92.5 & 87.3 & 93.1 & 90.1 & 93.6 & 88.6 & 95.4 & 92.5 & 91.9 & 90.9 & 92.3 \\
& + LSI & 94.4 & 92.9 & 92.7 & 90.8 & 94.5 & 86.1 & 90.3 & 89.9 & 92.8 & 88.5 & 94.4 & 91.6 & 90.1 & 90.5 & 91.4  \\
& + \neuron & 94.4 & 91.1 & 91.5 & 96.5 & 93.0 & 88.7 & 95.8 & 90.7 & 97.9 & 90.6 & 91.8 & 91.3 & 95.9 & 91.8 & 92.9\\
& + \vonename & 97.9 & 97.6 & 95.9 & 95.1 & 96.9 & 86.0 & 95.9 & 94.5 & 97.0 & 92.2 & 98.0 & 95.3 & 96.9 & \textbf{93.0} & 95.2 \\
% & + \vonename & 92.9 & 95.1 & 94.5 & 92.6 & 96.2 & 85.1 & 92.6 & 89.8 & 91.8 & 89.2 & 95.7 & 93.7 & 94.4 & 89.4 & 92.4 \\
%& + v2 & 92.5 & 93.8 & 94.2 & 92.7 & 94.6 & 83.9 & 92.3 & 88.5 & 91.9 & 91.3 & 95.8 & 94.7 & 90.8 & 89.2 & 91.9 \\
& + \mlpname & \textbf{98.9} & \textbf{99.3} & \textbf{98.3} & \textbf{97.6} & \textbf{98.3} & \textbf{91.1} & \cellcolor{cellground}\textbf{98.6} & \textbf{99.7} & \cellcolor{cellground}\textbf{98.6} & \textbf{92.8} & \textbf{98.2} & \cellcolor{cellground}\textbf{97.6} & \cellcolor{cellground}\textbf{97.6} & 91.3 & \textbf{97.0} \\
%98.6 & 98.6 & 98.3 & 99.7 & 98.3 & 92.4 & 97.2 & 95.2 & 96.8 & 94.5 & 99.3 & 98.3 & 96.3 & 90.1 & \textbf{96.7} \\% 99.3 & 99.7 & 98.6 & 98.6 & 99.7 & 94.2 & \cellcolor{cellground}94.1 & 97.3 & \cellcolor{cellground}96.7 & 94.5 & 99.3 & \cellcolor{cellground}97.2 & \cellcolor{cellground}95.8 & 94.6 & \textbf{97.1}  \\
\cmidrule{2-17}  \rowcolor{gray!10}
\cellcolor{white}& Gemma 2 & 87.0 & 91.0 & 91.0 & 87.0 & 84.0 & 71.0 & 89.0 & 76.0 & 77.0 & 91.0 & 77.0 & 83.0 & 71.0 & 70.0 & 81.8 \\
& + LSI & 58.1 & 78.0 & 86.2 & 55.4 & 84.7 & 76.4 & 77.2 & 81.2 & 76.2 & 88.8 & 93.2 & 88.8 & 94.2 & 77.0 & 80.0\\
& + \neuron & 71.2 & 70.9 & 81.6 & 76.9 & 75.8 & 64.8 & 71.8 & 77.9 & 83.2 & 81.0 & 78.0 & 71.7 & 73.0 & 72.8 & 75.3 \\
& + \vonename & 90.2 & 96.8 & 97.6 & 96.2 & \textbf{99.7} & 80.4 & 97.9 & 94.3 & 96.5 & \textbf{95.2} & 97.8 & 94.7 &\textbf{98.6} & 87.4 & 94.5 \\
% & + \vonename & 92.0 & 98.0 & 98.0 & 99.0 & 97.0 & 82.0 & 98.0 & 93.0 & 92.0 & 99.0 & 98.0 & 98.0 & 99.0 & 86.0 & \textbf{94.9} \\
%& + v2 & 86.6 & 100.0 & 98.0 & 98.0 & 100.0 & 80.0 & 100.0 & 94.0 & 100.0 & 96.0 & 100.0 & 99.0 & 100.0 & 95.0 & \textbf{96.2} \\
\cellcolor{white}\multirow{-18}{*}{\rotatebox[origin=c]{90}{zero-shot}} & + \mlpname & \textbf{94.4} & \textbf{98.6} & \textbf{100.0} & \textbf{98.6} & 99.6 & \textbf{89.8} & \cellcolor{cellground}\textbf{98.3} & \textbf{95.2} & \cellcolor{cellground}\textbf{99.3} & 94.2 & \textbf{98.9} & \cellcolor{cellground}\textbf{97.9} & \cellcolor{cellground}\textbf{98.6} &\textbf{88.8} & \textbf{96.6} \\ 
%93.3 & 98.9 & 99.0 & 98.6 & 99.6 & 86.8 & 98.6 & 95.5 & 99.6 & 94.5 & 99.3 & 97.2 & 97.9 & 87.1 & \textbf{96.1}\\
\midrule \rowcolor{gray!10}
\cellcolor{white}& LLama 3.1 & 94.6 & 96.6 & 95.9 & 94.6 & 97.6 & 86.8 & 96.6 & 93.6 & 94.8 & 92.9 & 96.6 & 97.6 & 96.6 & 90.2 & 94.3 \\
& + LSI & 99.3 & \textbf{99.7} & 96.2 & 97.6 & 96.4 & 91.2 & 99.0 & 95.0 & 97.4 & \textbf{97.3} & 99.0 & 96.7 & 97.8 & 93.6 & 96.9 \\
& + \vonename & \textbf{100.0} & 99.6 & \textbf{99.7} & \textbf{98.6} & \textbf{100.0} & 88.2 & \textbf{99.3} & \textbf{98.6} & \textbf{100.0} & 95.5 & 99.2 & \textbf{98.9} & 98.6 & \textbf{96.6} & \textbf{98.1} \\
 & + \mlpname & 98.5 & \textbf{99.7} & 98.3 & 98.3 & 99.3 & \textbf{92.1} & \cellcolor{cellground}98.3 & 98.3 & \cellcolor{cellground}\textbf{100.0} & 93.5 & \textbf{100.0} & \cellcolor{cellground}97.1 & \cellcolor{cellground}\textbf{99.7} & 92.2 & 97.5 \\
\cmidrule{2-17}  \rowcolor{gray!10}
\cellcolor{white}& Qwen 2.5 & 94.7 & 95.7 & 97.2 & 95.1 & 93.4 & 84.5 & 93.7 & 89.8 & 83.1 & 90.3 & 95.8 & 95.3 & 93.8 & 90.2 & 92.3 \\
& + LSI & 93.6 & 90.1 & 89.8 & 89.0 & 95.1 & 83.7 & 90.0 & 88.2 & 91.9 & 90.4 & 93.9 & 91.8 & 91.3 & 91.3 & 90.7 \\
& + \vonename & \textbf{100.0} & \textbf{99.3} & \textbf{99.0} & 98.3 & \textbf{99.7} & 90.0 & 98.3 & \textbf{99.0} & 99.6 & \textbf{97.0} & \textbf{99.7} & 96.9 & 98.6 & \textbf{95.3} & 97.9 \\
& + \mlpname & 99.7 & 98.9 & \textbf{99.0} & \textbf{99.0} & 99.6 & \textbf{92.1} &  \cellcolor{cellground}\textbf{99.3} & 97.9 &  \cellcolor{cellground}\textbf{100.0} & 96.3 & 99.6 &  \cellcolor{cellground}\textbf{97.8} &  \cellcolor{cellground}\textbf{100.0} & 95.2 & \textbf{98.2}\\
\cmidrule{2-17}  \rowcolor{gray!10}
\cellcolor{white} & Gemma 2 & 70.4 & 80.9 & 84.9 & 82.3 & 66.4 & 64.5 & 81.7 & 77.6 & 70.5 & 82.3 & 79.3 & 79.5 & 71.1 & 68.8 & 76.1 \\
& + LSI & 74.3 & 88.6 & 93.2 & 92.4 & 80.9 & 61.7 & 82.7 & 72.0 & 76.8 & 81.0 & 75.4 & 79.8 & 83.3 & 73.3 & 79.7 \\
& + \vonename & 96.2 & 98.6 & 96.4 & 94.0 & 98.6 & 77.0 & 95.5 & 92.9 & 97.1 & \textbf{96.2} & \textbf{100.0} & 95.5 & \textbf{98.0} & 90.9 & 94.8 \\
\multirow{-15}{*}{\rotatebox[origin=c]{90}{5-shot}} & + \mlpname & \textbf{98.5} & \textbf{99.3} & \textbf{98.6} & \textbf{96.3} & \textbf{100.0} & \textbf{91.7} & \cellcolor{cellground}\textbf{98.6} & \textbf{97.2} & \cellcolor{cellground}\textbf{99.6} & 93.5 & 99.2 & \cellcolor{cellground}\textbf{98.9} & \cellcolor{cellground}97.5 & \textbf{92.9} & \textbf{97.3}\\

\bottomrule
\end{tabular}}
\vspace{-1mm}
\caption{\crosslc results on the LCB. For our learned steering function (\mlpname), languages not seen during training are highlighted in (darker) grey.}
\label{tab:lcb_cross}
\vspace{-1.5mm}
\end{table*}

\begin{table*}[t]
\centering
\def\arraystretch{0.83}
\fontsize{8.2pt}{8.1pt}\selectfont
\resizebox{\linewidth}{!}{
%{
\begin{tabular}{lrrrrrrrrrrrrrrrrrr}\toprule
 &\multicolumn{2}{c}{de} &  \multicolumn{2}{c}{en} & 	\multicolumn{2}{c}{eu}  & \multicolumn{2}{c}{fa} & \multicolumn{2}{c}{fr} & \multicolumn{2}{c}{sw}	& \multicolumn{2}{c}{tr} & 	\multicolumn{2}{c}{zh} &  \multicolumn{2}{c}{AVG} \\
Model & LPR & Acc & LPR & Acc & LPR & Acc &LPR & Acc &LPR & Acc &LPR & Acc &LPR & Acc &LPR & Acc &LPR & Acc   \\
\midrule \rowcolor{gray!10}
LLama 3.1 & 36.7 & 22.1 & 11.9 & 0.9 & 48.1 & 16.2 & 34.9 & 13.7 & 32.7 & 23.7 & 38.6 & 22.2 & 36.7 & 17.1 & 25.1 & 16.9 & 33.1 & 16.6
\\
%\midrule
+ LSI &  19.9 & 25.3 & 61.4 & 16.3 & 26.1 & 7.6 & 34.1 & 4.7 & 23.7 & 24.7 & 47.2 & 9.1 & 34.2 & 8.3 & 20.7 & 14.3 & 33.4 & 13.8\\
+ \vonename & 96.8 & 34.9 & \textbf{98.7} & 34.9 & 92.1 & 13.7 & \textbf{98.9} & 22.7 & \textbf{97.2} & 27.9 & \textbf{72.4} & 13.1 & \textbf{96.5} & 21.3 & \textbf{96.6} & 34.9 & \textbf{93.7} & 25.4 \\
% + \vonename & 88.7 & 41.6 & 25.3 & 9.1 & 91.5 & 18.8 & 83.6 & 32.2 & 79.2 & 47.9 & 56.0 & 32.3 & 86.6 & 29.6 & 65.4 & 32.3 & \textbf{72.0} & \textbf{30.5} \\
% + v2 & 68.7 & 42.6 & 7.3 & 3.8 & 48.6 & 18.8 & 45.5 & 21.3 & 59.0 & 38.9 & 44.7 & 25.8 & 74.3 & 32.4 & 39.0 & 25.4 & 48.4 & 26.1 \\
+ \mlpname & \textbf{97.1} & \textbf{51.1} & 79.5 & \textbf{53.1} & \cellcolor{cellground}\textbf{94.0} & \cellcolor{cellground}\textbf{23.9} & 97.3 & \textbf{37.4} & 95.6 & \textbf{50.0} & \cellcolor{cellground}69.4 & \cellcolor{cellground}\textbf{25.8} & \cellcolor{cellground}95.0 & \cellcolor{cellground}\textbf{35.6} & 85.5 & \textbf{47.6} & 91.4 & \textbf{40.6}\\ 
% 97.0 & 36.8 & 96.4 & 39.2 & 89.8 & 17.8 & 98.6 & 21.8 & 92.2 & 39.5 & 63.3 & 19.2 & 96.5 & 26.9 & 86.9 & 30.7 & 90.1 & \textbf{29.0}\\ 
\midrule \rowcolor{gray!10}
Qwen 2.5 & 54.9 & 33.7 & \textbf{94.8} & \textbf{60.8} & 48.0 & 8.6 & 55.9 & 19.4 & 53.5 & 40.0 & 44.1 & 7.1 & 54.2 & 17.1 & 58.8 & 35.9 & 58.0 & 27.8 \\
%\midrule
+ LSI & 47.6 & 27.4 & 92.3 & 55.5 & 42.7 & 5.1 & 54.7 & 12.3 & 55.3 & 31.1 & 39.6 & 5.1 & 46.8 & 15.7 & 48.6 & 38.1 & 53.4 & 23.8 \\
+ \vonename & 94.6 & 34.2 & 93.2& 57.9 & 81.1 & \textbf{13.2} & 97.2 & \textbf{14.7} & 94.5 & 42.6 & 60.0 & \textbf{13.6} & 92.7 & \textbf{28.7} & 91.4 & \textbf{51.9} & 88.1 & 32.1 \\
% + \vonename & 69.1 & 40.5 & 94.7 & 58.4 & 55.9 & 7.1 & 72.2 & 20.9 & 76.0 & 44.2 & 49.0 & 8.1 & 78.2 & 19.9 & 61.8 & 40.2 & \textbf{69.6} & \textbf{29.9}\\
% + v2 & 65.3 & 40.5 & 94.3 & 59.8 & 50.0 & 8.1 & 64.6 & 20.9 & 68.2 & 39.5 & 46.0 & 9.6 & 66.4 & 20.4 & 63.7 & 40.7 & 64.8 & 29.9 \\
+ \mlpname & \textbf{97.3} & \textbf{44.2} & 93.2 & 57.9 & \cellcolor{cellground}\textbf{83.3} & \cellcolor{cellground}12.7 & \textbf{98.2} & 11.4 & \textbf{95.5} & \textbf{50.0} & \cellcolor{cellground}\textbf{72.6} & \cellcolor{cellground} 9.6 & \cellcolor{cellground}\textbf{95.4} & \cellcolor{cellground}27.3 & \textbf{91.5} & 45.5 & \textbf{90.9} & \textbf{32.3} \\
% 79.8 & 37.9 & 95.9 & 53.1 & 64.2 & 9.1 & 81.2 & 12.3 & 87.7 & 45.3 & 58.7 & 8.1 & 72.3 & 22.7 & 61.3 & 41.8 & 75.1 & 28.8\\
\midrule \rowcolor{gray!10}
Gemma 2 & 28.1 & 12.6 & \textbf{98.1} & 48.8 & 19.3 & 4.6 & 18.2 & 3.8 & 32.6 & 18.9 & 10.6 & 13.6 & 44.6 & 19.9 & 34.0 & 20.6 & 35.7 & 17.9
 \\
+ LSI & 64.9 & 18.4 & 94.2 & 36.8 & 14.0 & 1.5 & 4.8 & 0.0 & 61.2 & 22.1 & 21.2 & 6.1 & 47.3 & 14.8 & 20.0 & 14.3 & 41.0 & 14.3 \\
+ \vonename & 83.8 & 23.2 & 98.0 & 48.3 & 53.5 & \textbf{15.2} & 85.9 & 10.9 & 78.6 & 32.1 & 43.9 & \textbf{18.7} & \textbf{94.3} & \textbf{35.6} & 71.0 & 35.4 & 76.1 & 27.4 \\
% + \vonename & 76.4 & 21.6 & 96.8 & 46.4 & 24.2 & 6.6 & 48.9 & 7.6 & 67.5 & 27.4 & 24.0 & 11.6 & 90.0 & 31.9 & 44.7 & 27.0 & \textbf{59.1} & \textbf{22.5}\\
% + v2 & 92.9 & 19.5 & 98.0 & 41.6 & 74.7 & 15.2 & 93.2 & 8.1 & 83.4 & 26.8 & 53.9 & 16.2 & 94.2 & 26.4 & 73.4 & 28.0 & \textbf{83.0} & \textbf{22.7} \\
+ \mlpname & \textbf{96.2} & \textbf{42.1} & 98.0 & \textbf{53.1} & \cellcolor{cellground}\textbf{58.5} & \cellcolor{cellground}7.6 & \textbf{90.2} &\textbf{14.2} & \textbf{92.1} & \textbf{39.5} & \cellcolor{cellground}\textbf{55.8} & \cellcolor{cellground}16.7 & \cellcolor{cellground}93.8 & \cellcolor{cellground}24.5 & \textbf{79.1} & \textbf{37.6} & \textbf{83.0} & \textbf{29.4}\\
%92.2 & 34.7 & 97.4 & 52.2 & 54.0 & 5.6 & 89.1 & 13.7 & 89.2 & 34.7 & 52.1 & 15.7 & 96.4 & 25.9 & 79.9 & 38.1 & \textbf{81.3} & \textbf{27.6}\\
\bottomrule
\end{tabular}}
\vspace{-1mm}
\caption{\crosslc (LPR) and QA results (Acc) on MultiQ. Languages not seen in ReCoVeR+ training are highlighted in (darker) grey.}
\label{tab:multiq_cross}
\vspace{-1.5mm}
\end{table*}

\sparagraph{Monolingual Language Confusion}
The \monolc (i.e., where LLMs are expected to answer in the language of the prompt) results on LCB and MultiQ are shown in Table \ref{tab:mono_agg} (for detailed, per-language results see Tables \ref{tab:lcb_mono}, \ref{tab:multiq_mono}, and \ref{tab:lcb_wpr}).
%
%In the monolingual setup, the target language is implicitly specified through the language of the prompt. If there is no contradicting instruction in the prompt, the model should answer in the language of the prompt. The scores on the LCB and the MultiQ datasets are given in  for LCB and Table \ref{tab:multiq_mono} for MultiQ). 
%
On LCB, Llama and Qwen exhibit robust out-of-the-box performance (>98\% LPR), offering limited opportunity for further gains. Gemma's performance is notably weaker: it more often defaults to English, regardless of the prompt language. 
%, underscoring that, while certain models adeptly generate answeras i, 
%as 
%%
Steering with LSI \cite{yunfan-etal-2025-mitigating} reduces language confusion for all models on both LCB and MultiQ, but at the same time results in large drops of QA accuracy on MultiQ (from -4\,percentage points (pp) for Gemma to almost -12\,pp for Llama). Our \name variants overall mitigate language confusion comparably or better (e.g., +8pp compared to LSI for Gemma on LCB), but in contrast to LSI, our \name(+) language steering actually improves the QA performance for Qwen and Gemma. The QA performance gains are particularly prominent with our trained language steering, \nametrain (+5pp for Qwen and +9pp for Gemma). These results suggest that our language representations are not just effective on their own (\name), but also---considering that we train \nametrain on merely 4400 instances (see \ref{sec:expsetup})---that they enable highly sample-efficient learning of the language steering intervention.           

%the  diminishes accuracy on correctly answered items (in terms of both language and content). In contrast to that, the \nametrain method notably improves Gemma's and Qwen's monolingual language confusion and performance. \todo{How to address Llamas poor MultiQ results :(}

\rparagraph{Effect on Task Performance} When steering a model toward a specific language, it is essential to preserve task performance. To assess the impact of language steering, we evaluate MMLU \cite{hendrycks2021Measuring} translated into German, Spanish, Indonesian, Swahili, and Chinese using Llama 3.1. The model is prompted to output only the correct answer option. As shown in Table \ref{tab:mmlu}, \name maintains performance across languages: the absolute difference between the baseline (out-of-the-box) model and the steered model remains within 0.4 pp. With \name we do not sacrifice task performance to improve language confusion.

%\label{sec:crosslingual_results}

\begin{figure}[t]
    \centering
    \includegraphics[width=0.87\linewidth]{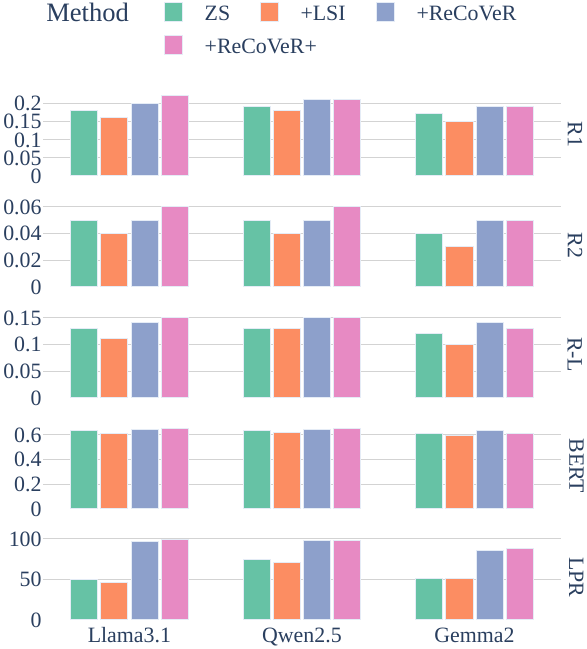}
    \vspace{-1mm}
    \caption{\crosslc (metric: LPR) and summarization performance (Rouge-1: R1; Rouge-2: R2; Rouge-L: R-L; and BERT-Score: BERT) on the CrossSum dataset for the language pairs: en-es, es-fr, fr-tr, and tr-sw.}
    \label{fig:crosssum_plot}
    \vspace{-1.5mm}
\end{figure}

\rparagraph{Crosslingual Language Confusion}
\crosslc results (LLM instructed to reply in a concrete language, different from the prompt language) on LCB, MultiQ, and CrossSum are shown in Table~\ref{tab:lcb_cross}, Table~\ref{tab:lcb_wpr}, Table~\ref{tab:multiq_cross}, and Figure~\ref{fig:crosssum_plot} (per language-pair results in Table~\ref{tab:cross_sum}), respectively. Owing to (1) computation of language vectors that is agnostic to any reference language (i.e., English) and (2) computation of the steering vectors using both representation $r_\mathit{source}$ of the prompt language and $r_\mathit{target}$ of the requested answer language, both \name variants excel in \crosslc: they massively outperform LSI across the board, for all three benchmarks, in zero-shot and few-shot evaluation and all three LLMs. \name variants again outperform LSI most prominently for Gemma, but the gains are often large for the other two LLMs too: e.g., on CrossSum, \name(+) yields +28pp over LSI (+24pp over the original model) for Qwen and +50pp (+46pp) for Llama. We also report the performance of language-specific neurons on LCB. Across all three models \name(+) outperforms language-specific neurons. \name consistently maintains or improves WPR. The additional training of \nametrain tends to improve the WPR over \name.

We note that all three models exhibit dramatically worse \crosslc performance on MultiQ (Table \ref{tab:multiq_cross}) than on the LCB dataset (Table \ref{tab:lcb_cross}), e.g., 91\% vs. 33\% for Llama. This is because the prompt language in MultiQ varies and is not fixed to English as in LCB. This shows that multilingual LLMs fail to comply with the requested response language much more often if the prompt is not in English. LSI actually consistently exhibits more \crosslc than the corresponding base models on MultiQ and CrossSum (Figure \ref{fig:crosssum_plot}) and this is also due to the non-English prompts in those datasets. LSI computes its language vectors aiming to mitigate answering in the dominant language. However, for non-English prompts, the model is prone to answer in the prompt language rather than English, which seems to render LSI inutile for cross-lingual applications with non-English prompts.

Figure~\ref{fig:examples} shows example generations of Llama without and with \name. Out-of-the-box Llama struggles to follow the instruction and reply in the correct language. Instead it replies in the source language. In the second example the content of the answer changes as well. ReCoVeR provides the correct answer but the sentence, while understandable, is not in correct grammar. 

In contrast to \monolc results, in \crosslc the unsupervised steering based on simple subtraction of language vectors (\name) almost matches the performance of the learned steering (\nametrain): this suggests that the difference $r_\mathit{source}-r_\mathit{target}$ already represents a very good steering vector for \crosslc and that the learned steering does not capture much more than this difference either.

\rparagraph{Generalization to Unseen Languages} On both LCB and MultiQ, we evaluate \nametrain on several target languages to which the model was not exposed during training of the steering intervention. The performance for those languages thus quantifies the extent of cross-lingual generalization of our learned steering function.   
%%%
%to which extent   We evaluated We intentionally exclude languages from the training set of the learned steering vectors. This enables us to explore how well \nametrain extends to languages not covered by the training set. 
The results in Table \ref{tab:lcb_cross} (languages: it, ko, tr, vi) and Table \ref{tab:multiq_cross} (languages: eu, sw, tr) indicate a successful cross-lingual of the mitigation of language confusion (i.e., \nametrain consistently matches or surpasses the performance of \name). The same, however, does not consistently hold for task-specific performance on MultiQ: especially for Gemma, transfer of the learned steering function of \nametrain to an unseen language can lead to a substantial performance drop (e.g., -8pp for Basque or -11pp for Turkish) compared to the unsupervised steering (\name). Although it is worth noting that even in these cases \nametrain almost always yields higher QA accuracy than the base model and steering with LSI.        

%suggest that the language confusion for unseen languages tends to improve. However, this does not necessarily extend to the downstream performance. The transfer to unseen languages seems to be especially challenging for Gemma 2, where performance drops by $11.1$ for Turkish questions. 

\subsection{Further Analyses}

\begin{figure}[t]
    \centering
    \includegraphics[width=0.87\linewidth]{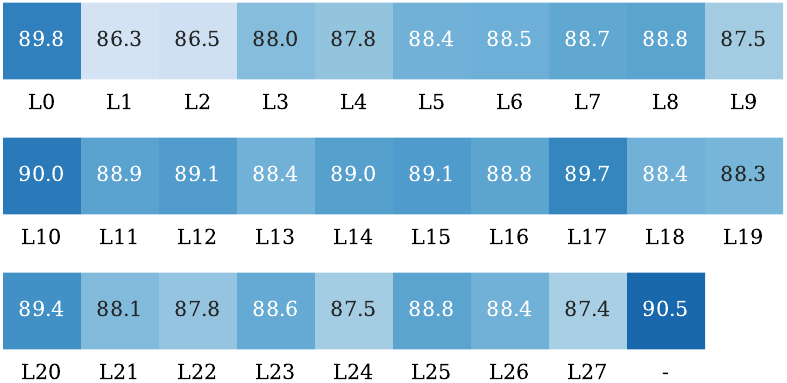}
    \caption{LPR for Qwen2.5 when we remove steering from individual layers.}
    \vspace{-1mm}
    \label{fig:leave_out}
    \vspace{-1.5mm}
\end{figure}

\sparagraph{Leave-Out Layers}
Previous work suggested that some layers produce representations that encode more language information than others \cite{bhattacharya-bojar-2023-unveiling,tang2024language}.
%not all layers of the model are language-specific. 
We thus next measure language confusion while omitting language steering in one transformer layer at a time. We run the experiment on Qwen, using the complex questions from LCB (languages: de, es, hi, ja, pt, zh). 
%skipping a layer of when applying the steering vectors. We use the complex prompts of the LCB benchmark in . 
Figure~\ref{fig:leave_out} summarizes the results: we achieve best performance ($90.5\%$ LPR) when applying steering in \textit{all} layers. 
%is achieved when all layers are considered for steering. 
We observe the largest performance drops if we remove the steering in layers 1 and 2: effective mitigation of language confusion seems to require very ``early'' steering. We again see larger drops if we remove steering from higher layers (L22-L27).
%This suggests that these layers are crucial to reducing language confusion. 
Our finding that language-specific steering is more important at the bottom and top layers than in middle layers is in line  
%The results suggest that the bottom and top layers are more important than the middle layers to reduce language confusion. This aligns 
with the hypothesis that the intermediate layers are responsible for reasoning, and thus rely on the English-centric representations \cite{zhao2024how}. 

\begin{figure}[t]
    \centering
    \includegraphics[width=0.87\linewidth]{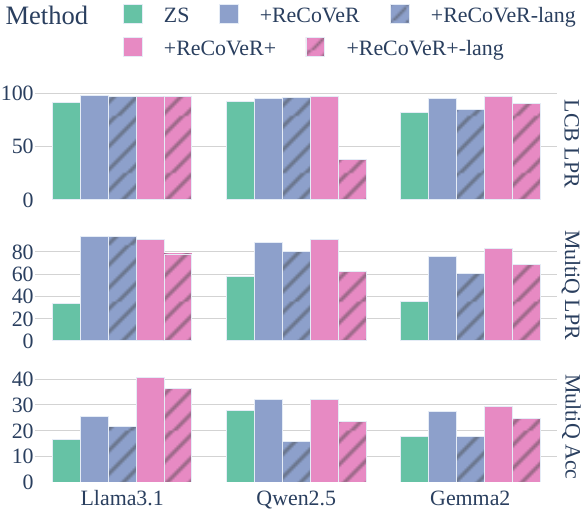}
    \caption{\crosslc performance on LCB and MultiQ without language specification in the prompt.}
    \label{fig:no_lang}
    \vspace{-1.5mm}
\end{figure}

\rparagraph{Steering as the Only Language Indication}
\label{sec:crosslingual_no_lang_results}
Prompts in the \crosslc evaluation explicitly specify the expected response language: \name steering then mitigates language confusion and helps LLMs generate text in the specified language.  
%in the generation. However, one interesting question is how strong the steering vectors are and whether they are sufficient to indicate the language.
We next investigate to which extent \name steering alone conditions the generation language, i.e., when we remove the answer language specification from the prompt. \name thus has to steer the model away from the implicitly specified prompt language and towards the target language: this requires stronger steering compared to prompts with explicit language specifications, and we find that optimal $\alpha=2$ for Qwen and $\alpha=4$ for Gemma. For Llama, however, $\alpha=1$ remains. The results, shown in Figure \ref{fig:no_lang}, reveal that, for Llama and Gemma, steering via \name(+) alone is more effective than steering by only specifying the response language in the prompt (ZS). 
%performs better than zero-shot performance with language information, but both language information with steering perform best. With 
%Moreover, for Llama, \nametrain steering alone is so effective that additionally specifying the expected response language within the prompt brings little additional performance gains. 
%language information is comparable to the performance with both steering and language information. Qwen, on the other hand, shows substantial performance drops. 

\begin{figure}[t!]
    \centering
    \includegraphics[width=0.83\linewidth]{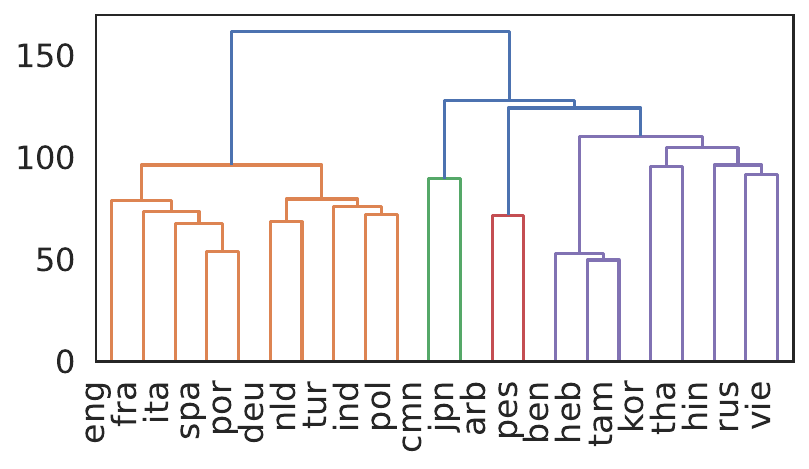}
    \caption{Visualization of the language representations of the last layer of Llama 3.1 using Agglomerative Hierarchical Clustering.}
    \label{fig:representation_pca}
    \vspace{-1mm}
\end{figure}

\rparagraph{Language Vectors} We next qualitatively analyze language vectors $r_l$ by grouping them via hierarchical agglomerative clustering, to see if they reflect known relations between languages (e.g., typology/genealogy, shared script and/or shared vocabulary). 
%To better understand the language representations, we visualize the representations of the last layer for all models using a Dendogram of Agglomerative Hierarchical Clustering in 5
Figure \ref{fig:representation_pca} displays the resulting dendrogram for Llama (Figure \ref{fig:representation_pca_appendix} shows Qwen and Gemma). 
%We would expect that the similarity of the representations correlates with language families, shared scripts, and shared vocabulary. 
Our language steers capture, to some extent, language families: all models generally group together Indo-European Languages and in particular the four Romance languages (French, Spanish, Italian and Portuguese; with Spanish and Portuguese having consistently most similar vectors). 
%, all in the Italic branch of the Indo-European Languages, are especially close in the representation space. 
There is some evidence that a shared script drives the proximity: (1) representations from all three models tend to group Latin-script languages; and (2) Llama vectors put Chinese and Japanese close together: although Japanese is in a different language family and also typologically dissimilar to Chinese, it borrows Chinese scripts and a non-negligible portion of Chinese vocabulary.  % although they are from different families and are t.   similarly. These two languages are from separate language families, but Japanese also includes Chinese scripts and some Chinese vocabulary, justifying similar representations. 
Qwen, on the other hand, represents Chinese similar to Hindi, Bengali, and Tamil. These languages are from different language families, and generally have little in common other than geographic proximity.
%, they have little in common. 

%The clusters also suggest that languages with Latin scripts have similar representations. Qwen clusters languages with Latin scripts and languages with other scripts. Llama and Gemma cluster all Latin-script languages except Vietnamese together.
%%%%%%%%%
%While there are some parallels to the language families and languages in Latin script tend to be close, not all languages close in the representation space share similarities.

\rparagraph{Negative Result: Position-Specific Steering}
In \name, we compute language-specific vectors at the level of layers, i.e., we steer representations of \textit{all} tokens in a layer with the same steering vector. Initially, however, we intended to compute/learn position-specific steering vectors, i.e., a different steering vector for each layer and each token position. However, positional information largely gets lost due to averaging across samples: because of varying sentence structures and sequence lengths, different positions end up with very similar representations. Because of this, position-specific steering consistently produced worse results than layer-level steering (we report the results in \S\ref{app:positions}).       
% and language, capturing how the model internally encodes that language in its hidden states. In the following, we will explore these representations.

%%%
%Initially, we hypothesized that mitigating language confusion with language-specific steering vectors would benefit from positional information. Thus, we explored using language vectors $\hat{v}_{l}$ that average the hidden states per position. 
%However, computing $\hat{v}_{l}$ showed that the positional information gets lost during averaging across samples. Probably due to varying sentence structures and sequence length, the different positions have very similar representations. To address different sequence lengths, we also experiment with splitting the sequence into $k$ chunks and computing the average per chunk. This maintains information about the position in the text, e.g. start, middle, or end of text. However, this exhibits the same problem as the representations for each position. 
%Thus, computing the positional language vectors has little to no advantage over the language vectors $v_{l}$ averaged across positions. In fact, they tend to reduce performance as shown in Table \ref{tab:positional}.

%Future work could explore structuring the sentences, for instance, with dependency parsing to obtain representations that encapsulate positional information. 

\section{Conclusion}
We introduced \name, a novel lightweight language steering approach. We first isolate language vectors using a multi-parallel corpus and then leverage those vectors for effective language steering of LLMs via (i) unsupervised steering arithmetic as well as (ii) learnable steering intervention, trained in a sample-efficient manner. Our extensive evaluation, encompassing three benchmarks and 18 languages, shows that, in contrast to prior approaches, \name effectively mitigates language confusion without jeopardizing task performance. Future work could leverage linguistic information to obtain more effective language steering, e.g., vectors conditioned by (i) syntactic roles, exploiting multilingual dependency parsers \cite{de2021universal} or (ii) typological features, e.g., URIEL \cite{littell2017uriel}, to facilitate cross-lingual generalization (i.e., transfer) for learning steering vectors.  

%Future work could explore structuring the sentences, for instance, with dependency parsing to obtain representations that encapsulate positional information. 

\section*{Limitations}
Our experiments focus on a broad range of languages across different families, offering wide-ranging initial insights. While this provides substantial coverage, the applicability of the current results to the entirety of global languages requires further investigation. 

We evaluate \name on a diverse set of state-of-the-art multilingual model families, including Llama, Qwen, and Gemma. These represent widely adopted multilingual models, providing a strong basis for evaluating the generalizability of \name. The results suggest that \name is likely to generalize well beyond the models tested.

We use FLORES-200 as the multi-parallel data source to compute language representations. Although it provides meaningful language representations, demonstrating that our findings extend to other multi-parallel datasets requires further experiments. Moreover, we use all available samples in FLORES-200, and thus do not address the question of how many samples are minimally required to obtain meaningful language representations.

\section*{Acknowledgements}
Hannah Sterz thanks the Cambridge Trust for their support
via the International Scholarship. This work has been supported by a Royal Society University Research Fellowship \textit{‘Inclusive and Sustainable Language Technology for a Truly Multilingual World’} (no 221137) awarded to Ivan Vuli\'{c}.

\bibliography{anthology,custom}

\begin{thebibliography}{37}
\providecommand{\natexlab}[1]{#1}

\bibitem[{Alain and Bengio(2017)}]{alain2017understanding}
Guillaume Alain and Yoshua Bengio. 2017.
\newblock \href {https://openreview.net/forum?id=HJ4-rAVtl} {Understanding intermediate layers using linear classifier probes}.
\newblock In \emph{5th International Conference on Learning Representations, {ICLR} 2017, Toulon, France, April 24-26, 2017, Workshop Track Proceedings}. OpenReview.net.

\bibitem[{Aryabumi et~al.(2024)Aryabumi, Dang, Talupuru, Dash, Cairuz, Lin, Venkitesh, Smith, Campos, Tan et~al.}]{aryabumi2024aya}
Viraat Aryabumi, John Dang, Dwarak Talupuru, Saurabh Dash, David Cairuz, Hangyu Lin, Bharat Venkitesh, Madeline Smith, Jon~Ander Campos, Yi~Chern Tan, and 1 others. 2024.
\newblock Aya 23: Open weight releases to further multilingual progress.
\newblock \emph{arXiv preprint arXiv:2405.15032}.

\bibitem[{Bhattacharjee et~al.(2023)Bhattacharjee, Hasan, Ahmad, Li, Kang, and Shahriyar}]{bhattacharjee-etal-2023-crosssum}
Abhik Bhattacharjee, Tahmid Hasan, Wasi~Uddin Ahmad, Yuan-Fang Li, Yong-Bin Kang, and Rifat Shahriyar. 2023.
\newblock \href {https://doi.org/10.18653/v1/2023.acl-long.143} {{C}ross{S}um: Beyond {E}nglish-centric cross-lingual summarization for 1,500+ language pairs}.
\newblock In \emph{Proceedings of the 61st Annual Meeting of the Association for Computational Linguistics (Volume 1: Long Papers)}, pages 2541--2564, Toronto, Canada. Association for Computational Linguistics.

\bibitem[{Bhattacharya and Bojar(2023)}]{bhattacharya-bojar-2023-unveiling}
Sunit Bhattacharya and Ond{\v{r}}ej Bojar. 2023.
\newblock \href {https://doi.org/10.18653/v1/2023.blackboxnlp-1.9} {Unveiling multilinguality in transformer models: Exploring language specificity in feed-forward networks}.
\newblock In \emph{Proceedings of the 6th BlackboxNLP Workshop: Analyzing and Interpreting Neural Networks for NLP}, pages 120--126, Singapore. Association for Computational Linguistics.

\bibitem[{Bricken et~al.(2023)Bricken, Templeton, Batson, Chen, Jermyn, Conerly, Turner, Anil, Denison, Askell et~al.}]{bricken2023towards}
Trenton Bricken, Adly Templeton, Joshua Batson, Brian Chen, Adam Jermyn, Tom Conerly, Nick Turner, Cem Anil, Carson Denison, Amanda Askell, and 1 others. 2023.
\newblock Towards monosemanticity: Decomposing language models with dictionary learning.
\newblock \emph{Transformer Circuits Thread}, 2.

\bibitem[{Cao et~al.(2024)Cao, Zhang, Cao, Yin, Lin, Ma, and Chen}]{cao2024personalized}
Yuanpu Cao, Tianrong Zhang, Bochuan Cao, Ziyi Yin, Lu~Lin, Fenglong Ma, and Jinghui Chen. 2024.
\newblock \href {https://openreview.net/forum?id=7qJFkuZdYo} {Personalized steering of large language models: Versatile steering vectors through bi-directional preference optimization}.
\newblock In \emph{The Thirty-eighth Annual Conference on Neural Information Processing Systems}.

\bibitem[{Costa-Juss{\`a} et~al.(2022)Costa-Juss{\`a}, Cross, {\c{C}}elebi, Elbayad, Heafield, Heffernan, Kalbassi, Lam, Licht, Maillard et~al.}]{costa2022no}
Marta~R Costa-Juss{\`a}, James Cross, Onur {\c{C}}elebi, Maha Elbayad, Kenneth Heafield, Kevin Heffernan, Elahe Kalbassi, Janice Lam, Daniel Licht, Jean Maillard, and 1 others. 2022.
\newblock No language left behind: Scaling human-centered machine translation.
\newblock \emph{arXiv preprint arXiv:2207.04672}.

\bibitem[{De~Marneffe et~al.(2021)De~Marneffe, Manning, Nivre, and Zeman}]{de2021universal}
Marie-Catherine De~Marneffe, Christopher~D Manning, Joakim Nivre, and Daniel Zeman. 2021.
\newblock Universal dependencies.
\newblock \emph{Computational linguistics}, 47(2):255--308.

\bibitem[{Grattafiori et~al.(2024)Grattafiori, Dubey, Jauhri, Pandey, Kadian, Al-Dahle, Letman, Mathur, Schelten, Vaughan et~al.}]{grattafiori2024llama}
Aaron Grattafiori, Abhimanyu Dubey, Abhinav Jauhri, Abhinav Pandey, Abhishek Kadian, Ahmad Al-Dahle, Aiesha Letman, Akhil Mathur, Alan Schelten, Alex Vaughan, and 1 others. 2024.
\newblock The llama 3 herd of models.
\newblock \emph{arXiv preprint arXiv:2407.21783}.

\bibitem[{Hendrycks et~al.(2021)Hendrycks, Burns, Basart, Zou, Mazeika, Song, and Steinhardt}]{hendrycks2021Measuring}
Dan Hendrycks, Collin Burns, Steven Basart, Andy Zou, Mantas Mazeika, Dawn Song, and Jacob Steinhardt. 2021.
\newblock \href {https://openreview.net/forum?id=d7KBjmI3GmQ} {Measuring massive multitask language understanding}.
\newblock In \emph{9th International Conference on Learning Representations, {ICLR} 2021, Virtual Event, Austria, May 3-7, 2021}. OpenReview.net.

\bibitem[{Hernandez et~al.(2024)Hernandez, Li, and Andreas}]{hernandez2024inspecting}
Evan Hernandez, Belinda~Z. Li, and Jacob Andreas. 2024.
\newblock \href {https://openreview.net/forum?id=ADtL6fgNRv} {Inspecting and editing knowledge representations in language models}.
\newblock In \emph{First Conference on Language Modeling}.

\bibitem[{Holtermann et~al.(2024)Holtermann, R{\"o}ttger, Dill, and Lauscher}]{holtermann-etal-2024-evaluating}
Carolin Holtermann, Paul R{\"o}ttger, Timm Dill, and Anne Lauscher. 2024.
\newblock \href {https://doi.org/10.18653/v1/2024.findings-acl.265} {Evaluating the elementary multilingual capabilities of large language models with {M}ulti{Q}}.
\newblock In \emph{Findings of the Association for Computational Linguistics: ACL 2024}, pages 4476--4494, Bangkok, Thailand. Association for Computational Linguistics.

\bibitem[{Jorgensen et~al.(2024)Jorgensen, Cope, Schoots, and Shanahan}]{jorgensen2023improving}
Ole Jorgensen, Dylan Cope, Nandi Schoots, and Murray Shanahan. 2024.
\newblock Improving activation steering in language models with mean-centring.
\newblock In \emph{Responsible Language Models Workshop at AAAI-24}.

\bibitem[{Kojima et~al.(2024)Kojima, Okimura, Iwasawa, Yanaka, and Matsuo}]{kojima-etal-2024-multilingual}
Takeshi Kojima, Itsuki Okimura, Yusuke Iwasawa, Hitomi Yanaka, and Yutaka Matsuo. 2024.
\newblock \href {https://doi.org/10.18653/v1/2024.naacl-long.384} {On the multilingual ability of decoder-based pre-trained language models: Finding and controlling language-specific neurons}.
\newblock In \emph{Proceedings of the 2024 Conference of the North American Chapter of the Association for Computational Linguistics: Human Language Technologies (Volume 1: Long Papers)}, pages 6919--6971, Mexico City, Mexico. Association for Computational Linguistics.

\bibitem[{Ladhak et~al.(2020)Ladhak, Durmus, Cardie, and McKeown}]{ladhak-etal-2020-wikilingua}
Faisal Ladhak, Esin Durmus, Claire Cardie, and Kathleen McKeown. 2020.
\newblock \href {https://doi.org/10.18653/v1/2020.findings-emnlp.360} {{W}iki{L}ingua: A new benchmark dataset for cross-lingual abstractive summarization}.
\newblock In \emph{Findings of the Association for Computational Linguistics: EMNLP 2020}, pages 4034--4048, Online. Association for Computational Linguistics.

\bibitem[{Lambert et~al.(2024)Lambert, Morrison, Pyatkin, Huang, Ivison, Brahman, Miranda, Liu, Dziri, Lyu et~al.}]{lambert2024tulu}
Nathan Lambert, Jacob Morrison, Valentina Pyatkin, Shengyi Huang, Hamish Ivison, Faeze Brahman, Lester James~V Miranda, Alisa Liu, Nouha Dziri, Shane Lyu, and 1 others. 2024.
\newblock Tulu 3: Pushing frontiers in open language model post-training.
\newblock \emph{arXiv preprint arXiv:2411.15124}.

\bibitem[{Littell et~al.(2017)Littell, Mortensen, Lin, Kairis, Turner, and Levin}]{littell2017uriel}
Patrick Littell, David~R Mortensen, Ke~Lin, Katherine Kairis, Carlisle Turner, and Lori Levin. 2017.
\newblock Uriel and lang2vec: Representing languages as typological, geographical, and phylogenetic vectors.
\newblock In \emph{Proceedings of the 15th Conference of the European Chapter of the Association for Computational Linguistics: Volume 2, Short Papers}, pages 8--14.

\bibitem[{Liu et~al.(2024)Liu, Ye, Xing, and Zou}]{liu2023context}
Sheng Liu, Haotian Ye, Lei Xing, and James~Y. Zou. 2024.
\newblock \href {https://openreview.net/forum?id=dJTChKgv3a} {In-context vectors: Making in context learning more effective and controllable through latent space steering}.
\newblock In \emph{Forty-first International Conference on Machine Learning, {ICML} 2024, Vienna, Austria, July 21-27, 2024}. OpenReview.net.

\bibitem[{Marchisio et~al.(2024)Marchisio, Ko, Berard, Dehaze, and Ruder}]{marchisio-etal-2024-understanding}
Kelly Marchisio, Wei-Yin Ko, Alexandre Berard, Th{\'e}o Dehaze, and Sebastian Ruder. 2024.
\newblock \href {https://doi.org/10.18653/v1/2024.emnlp-main.380} {Understanding and mitigating language confusion in {LLM}s}.
\newblock In \emph{Proceedings of the 2024 Conference on Empirical Methods in Natural Language Processing}, pages 6653--6677, Miami, Florida, USA. Association for Computational Linguistics.

\bibitem[{Park et~al.(2024)Park, Choe, and Veitch}]{park2024linear}
Kiho Park, Yo~Joong Choe, and Victor Veitch. 2024.
\newblock The linear representation hypothesis and the geometry of large language models.
\newblock In \emph{International Conference on Machine Learning}, pages 39643--39666. PMLR.

\bibitem[{Rei et~al.(2022)Rei, Treviso, Guerreiro, Zerva, Farinha, Maroti, C.~de Souza, Glushkova, Alves, Coheur, Lavie, and Martins}]{rei-etal-2022-cometkiwi}
Ricardo Rei, Marcos Treviso, Nuno~M. Guerreiro, Chrysoula Zerva, Ana~C Farinha, Christine Maroti, Jos{\'e}~G. C.~de Souza, Taisiya Glushkova, Duarte Alves, Luisa Coheur, Alon Lavie, and Andr{\'e} F.~T. Martins. 2022.
\newblock \href {https://aclanthology.org/2022.wmt-1.60/} {{C}omet{K}iwi: {IST}-unbabel 2022 submission for the quality estimation shared task}.
\newblock In \emph{Proceedings of the Seventh Conference on Machine Translation (WMT)}, pages 634--645, Abu Dhabi, United Arab Emirates (Hybrid). Association for Computational Linguistics.

\bibitem[{Rimsky et~al.(2024)Rimsky, Gabrieli, Schulz, Tong, Hubinger, and Turner}]{rimsky-etal-2024-steering}
Nina Rimsky, Nick Gabrieli, Julian Schulz, Meg Tong, Evan Hubinger, and Alexander Turner. 2024.
\newblock \href {https://doi.org/10.18653/v1/2024.acl-long.828} {Steering llama 2 via contrastive activation addition}.
\newblock In \emph{Proceedings of the 62nd Annual Meeting of the Association for Computational Linguistics (Volume 1: Long Papers)}, pages 15504--15522, Bangkok, Thailand. Association for Computational Linguistics.

\bibitem[{Singh et~al.(2024)Singh, Ravfogel, Herzig, Aharoni, Cotterell, and Kumaraguru}]{singh2024representation}
Shashwat Singh, Shauli Ravfogel, Jonathan Herzig, Roee Aharoni, Ryan Cotterell, and Ponnurangam Kumaraguru. 2024.
\newblock \href {https://openreview.net/forum?id=GwA4go0Mw4} {Representation surgery: Theory and practice of affine steering}.
\newblock In \emph{Forty-first International Conference on Machine Learning, {ICML} 2024, Vienna, Austria, July 21-27, 2024}. OpenReview.net.

\bibitem[{Stoehr et~al.(2024)Stoehr, Du, Sn{\ae}bjarnarson, West, Cotterell, and Schein}]{stoehr-etal-2024-activation}
Niklas Stoehr, Kevin Du, V{\'e}steinn Sn{\ae}bjarnarson, Robert West, Ryan Cotterell, and Aaron Schein. 2024.
\newblock \href {https://doi.org/10.18653/v1/2024.findings-emnlp.479} {Activation scaling for steering and interpreting language models}.
\newblock In \emph{Findings of the Association for Computational Linguistics: EMNLP 2024}, pages 8189--8200, Miami, Florida, USA. Association for Computational Linguistics.

\bibitem[{Stolfo et~al.(2025)Stolfo, Balachandran, Yousefi, Horvitz, and Nushi}]{stolfo2024improving}
Alessandro Stolfo, Vidhisha Balachandran, Safoora Yousefi, Eric Horvitz, and Besmira Nushi. 2025.
\newblock \href {https://openreview.net/forum?id=wozhdnRCtw} {Improving instruction-following in language models through activation steering}.
\newblock In \emph{The Thirteenth International Conference on Learning Representations, {ICLR} 2025, Singapore, April 24-28, 2025}. OpenReview.net.

\bibitem[{Subramani et~al.(2022)Subramani, Suresh, and Peters}]{subramani-etal-2022-extracting}
Nishant Subramani, Nivedita Suresh, and Matthew Peters. 2022.
\newblock \href {https://doi.org/10.18653/v1/2022.findings-acl.48} {Extracting latent steering vectors from pretrained language models}.
\newblock In \emph{Findings of the Association for Computational Linguistics: ACL 2022}, pages 566--581, Dublin, Ireland. Association for Computational Linguistics.

\bibitem[{Tang et~al.(2024{\natexlab{a}})Tang, Luo, Huang, Zhang, Wang, Zhao, Wei, and Wen}]{tang2024language}
Tianyi Tang, Wenyang Luo, Haoyang Huang, Dongdong Zhang, Xiaolei Wang, Wayne~Xin Zhao, Furu Wei, and Ji-Rong Wen. 2024{\natexlab{a}}.
\newblock Language-specific neurons: The key to multilingual capabilities in large language models.
\newblock In \emph{Proceedings of the 62nd Annual Meeting of the Association for Computational Linguistics (Volume 1: Long Papers)}, pages 5701--5715.

\bibitem[{Tang et~al.(2024{\natexlab{b}})Tang, Luo, Huang, Zhang, Wang, Zhao, Wei, and Wen}]{tang-etal-2024-language}
Tianyi Tang, Wenyang Luo, Haoyang Huang, Dongdong Zhang, Xiaolei Wang, Xin Zhao, Furu Wei, and Ji-Rong Wen. 2024{\natexlab{b}}.
\newblock \href {https://doi.org/10.18653/v1/2024.acl-long.309} {Language-specific neurons: The key to multilingual capabilities in large language models}.
\newblock In \emph{Proceedings of the 62nd Annual Meeting of the Association for Computational Linguistics (Volume 1: Long Papers)}, pages 5701--5715, Bangkok, Thailand. Association for Computational Linguistics.

\bibitem[{Team et~al.(2025)Team, Kamath, Ferret, Pathak, Vieillard, Merhej, Perrin, Matejovicova, Ram{\'e}, Rivi{\`e}re et~al.}]{team2025gemma}
Gemma Team, Aishwarya Kamath, Johan Ferret, Shreya Pathak, Nino Vieillard, Ramona Merhej, Sarah Perrin, Tatiana Matejovicova, Alexandre Ram{\'e}, Morgane Rivi{\`e}re, and 1 others. 2025.
\newblock Gemma 3 technical report.
\newblock \emph{arXiv preprint arXiv:2503.19786}.

\bibitem[{Team et~al.(2024)Team, Riviere, Pathak, Sessa, Hardin, Bhupatiraju, Hussenot, Mesnard, Shahriari, Ram{\'e} et~al.}]{team2024gemma}
Gemma Team, Morgane Riviere, Shreya Pathak, Pier~Giuseppe Sessa, Cassidy Hardin, Surya Bhupatiraju, L{\'e}onard Hussenot, Thomas Mesnard, Bobak Shahriari, Alexandre Ram{\'e}, and 1 others. 2024.
\newblock Gemma 2: Improving open language models at a practical size.
\newblock \emph{arXiv preprint arXiv:2408.00118}.

\bibitem[{Turner et~al.(2023)Turner, Thiergart, Leech, Udell, Vazquez, Mini, and MacDiarmid}]{turner2023activation}
Alexander~Matt Turner, Lisa Thiergart, Gavin Leech, David Udell, Juan~J Vazquez, Ulisse Mini, and Monte MacDiarmid. 2023.
\newblock Activation addition: Steering language models without optimization.
\newblock \emph{arXiv e-prints}, pages arXiv--2308.

\bibitem[{Wang et~al.(2025)Wang, YANG, and Peng}]{wang2025semanticsadaptive}
Weixuan Wang, JINGYUAN YANG, and Wei Peng. 2025.
\newblock \href {https://openreview.net/forum?id=8WQ7VTfPTl} {Semantics-adaptive activation intervention for {LLM}s via dynamic steering vectors}.
\newblock In \emph{The Thirteenth International Conference on Learning Representations}.

\bibitem[{Yang et~al.(2024)Yang, Yang, Zhang, Hui, Zheng, Yu, Li, Liu, Huang, Wei et~al.}]{yang2024qwen2}
An~Yang, Baosong Yang, Beichen Zhang, Binyuan Hui, Bo~Zheng, Bowen Yu, Chengyuan Li, Dayiheng Liu, Fei Huang, Haoran Wei, and 1 others. 2024.
\newblock Qwen2. 5 technical report.
\newblock \emph{arXiv preprint arXiv:2412.15115}.

\bibitem[{Yunfan et~al.(2025)Yunfan, Zou, Luo, Tang, Li, Luo, and Dong}]{yunfan-etal-2025-mitigating}
Xie Yunfan, Lixin Zou, Dan Luo, Min Tang, Chenliang Li, Xiangyang Luo, and Liming Dong. 2025.
\newblock \href {https://aclanthology.org/2025.coling-main.563/} {Mitigating language confusion through inference-time intervention}.
\newblock In \emph{Proceedings of the 31st International Conference on Computational Linguistics}, pages 8418--8431, Abu Dhabi, UAE. Association for Computational Linguistics.

\bibitem[{Zhang and Viteri(2025)}]{zhang2025uncovering}
Jason Zhang and Scott~W Viteri. 2025.
\newblock \href {https://openreview.net/forum?id=ICuIdJzBPm} {Uncovering latent chain of thought vectors in large language models}.
\newblock In \emph{Workshop on Neural Network Weights as a New Data Modality}.

\bibitem[{Zhao et~al.(2024)Zhao, Zhang, Chen, Kawaguchi, and Bing}]{zhao2024how}
Yiran Zhao, Wenxuan Zhang, Guizhen Chen, Kenji Kawaguchi, and Lidong Bing. 2024.
\newblock \href {https://proceedings.neurips.cc/paper_files/paper/2024/file/1bd359b32ab8b2a6bbafa1ed2856cf40-Paper-Conference.pdf} {How do large language models handle multilingualism?}
\newblock In \emph{Advances in Neural Information Processing Systems}, volume~37, pages 15296--15319. Curran Associates, Inc.

\bibitem[{Zou et~al.(2023)Zou, Phan, Chen, Campbell, Guo, Ren, Pan, Yin, Mazeika, Dombrowski et~al.}]{zou2310representation}
Andy Zou, Long Phan, Sarah Chen, James Campbell, Phillip Guo, Richard Ren, Alexander Pan, Xuwang Yin, Mantas Mazeika, Ann-Kathrin Dombrowski, and 1 others. 2023.
\newblock Representation engineering: A top-down approach to ai transparency.
\newblock \emph{arXiv preprint arXiv:2310.01405}.

\end{thebibliography}
%\bibliography{custom}

\appendix

\section{Trainings and Evaluation Details}
\subsection{Hyperparameters}
\name depends on the choice of hyperparameters. Choosing suitable $\alpha$ and whether to restore the norm is crucial for reducing language confusion. We choose the hyperparameters based on the performance on a smaller dataset of 600 machine translated samples from the alpaca dataset. We measure LPR and exclude hyperparameters that result in unreadable text; e.g. by repeating a word or sequence of words corresponding to the language. The hyperparameters are listed in Table \ref{tab:hyperparams}. The monolingual scenario requires less steering and a smaller $\alpha$, except for Qwen.
\begin{table}[h!]
    \small
    \centering
    \begin{tabular}{lllccc}
        \toprule
        Model & Version & Task &  alpha & beta & norm  \\
        \midrule
         Llama 3.1 & \vonename & cross & 0.2 & - & true \\
          & \vonename & mono & 0.05 & - & true \\
          & \mlpname & both & 0.1 & 0.9 & true \\
          \midrule
        Qwen 2.5 & \vonename & cross & 1 & - & true \\
         & \vonename & mono & 2 & - & true \\
         & \mlpname & both & 1 & 0.9 & true \\
         \midrule
         Gemma 2 & \vonename & cross & 2 & - & false\\
         & \vonename & mono & 0.5 & - & true \\
         & \mlpname & both & 2 & 0.9 & true \\
       \bottomrule 
     \end{tabular}
    \caption{Hyperparameter used in the experiments, covering alpha, beta, and whether to restore the norm. }
    \label{tab:hyperparams}
   
\end{table}

LSI requires choosing hyperparameters $\tau, \gamma$. We list the hyperparameters in \ref{tab:hyperparams_lsi}. We base the search space on the range used in the original paper $\tau \in [0.02, 0.04, 0.06, 0.08, 0.1]$ and $\gamma \in [0.2, 0.4, 0.6, 0.8, 1.0]$. However, for Gemma 2 we observe a model collapse for parameters in this range. Therefore, we reduce $\gamma$ to $0.01$, generating coherent text in the target language. 
 
\begin{table}[h!]
    \small
    \centering
    \begin{tabular}{llcc}
        \toprule
        Model  & Task &  $\tau $ &  $\gamma$  \\
        \midrule
         Llama 3.1 & cross & 0.06 & 0.6 \\
           & mono &  0.04 & 0.6 \\
          \midrule
        Qwen 2.5  & cross & 0.06 & 0.2 \\
          & mono & 0.04 & 0.4 \\
         \midrule
         Gemma 2 & cross & 0.04 & 0.01\\
          & mono & 0.04 & 0.01  \\
       \bottomrule 
    \end{tabular}
    \caption{Hyperparameter used for LSI in the experiments}
    \label{tab:hyperparams_lsi}
\end{table}

\subsection{Learnable Steering Function: Training Details}
\label{sec:trdetails_recoverplus}

We train \nametrain with the following hyperparameters: we train for 1 epoch with a learning rate of $1e-6$ and Adam optimiser and $0.2$ dropout, and a rank of $32$. The model-specific hyperparameters are listed in Table \ref{tab:hyperparams}.

To isolate the influence of the language vectors on the generated text, the input prompts are formulated without explicit information specifying the target language. This ensures that the language selection in the output is guided by the learned steering vectors, rather than by direct cues within the prompts themselves.
% \section{Language Vectors}
% \subsection{Similarity of language vectors}
% The language vectors encode how the model represents languages internally. Ideally, the model also encodes the relation between languages: which language is close to another language in terms of language families, script vocabulary, ... We evaluate the cosine similarities between the language representations of the last layer of the model. They are illustrated in Figure \ref{fig:similarity}.

% For all three models, we observe a that some language families are clustered. Spanish, Portuguese, and Italian, which are all in the Italic branch of the Indo-European languages, have the highest similarity scores. In general, Indo-European languages tend to be more similar to each other than other languages. For non-European languages the clusters are not clear. Llama 3.1 represents Chinese and Japanese with similar vectors. These two languages are from separate language families, but Japanese also includes Chinese scripts and some Chinese vocabulary justifying similar representations. Qwen 2.5 and Gemma, on the other hand, represent Chinese and Hindi with similar directions. These two languages are from separate language families as well and, apart from their geographic proximity, they have little in common. 

\subsection{Dataset Details}
\label{app:dataset}

We create a multilingual instruction tuning dataset that contains monolingual (prompt and answer are in the same language) and crosslingual (prompt and answer are in different languages) samples. Table \ref{tab:dataset_stats} shows the number of samples per language for both. The dataset covers 18 languages from diverse language families and with varying scripts. 

Recognizing that state-of-the-art LLMs are more prone to language confusion in crosslingual versus monolingual generation, we have increased the proportion of crosslingual samples compared to monolingual samples for each language in our training data.

\begin{figure*}
    \centering
    \begin{subfigure}[b]{0.49\linewidth}
        \centering
        \includegraphics[width=\linewidth]{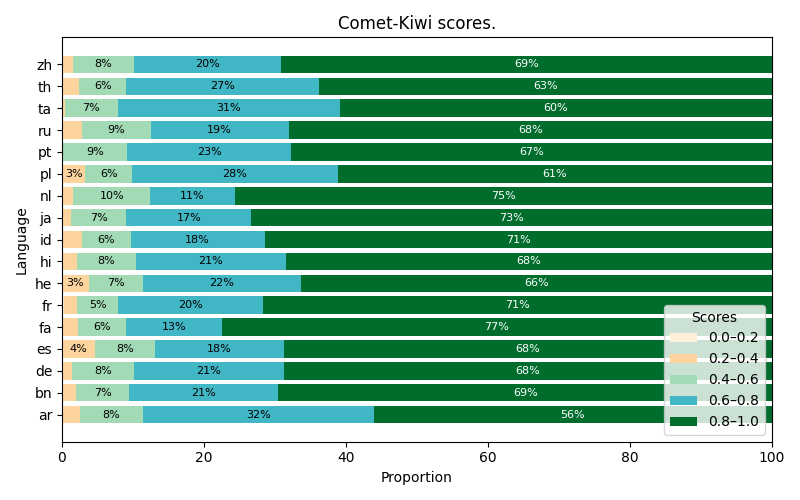}
        \caption{Comet-Kiwi Scores on the translated Questions.}
    \end{subfigure}
    \begin{subfigure}[b]{0.49\linewidth}
        \centering
        \includegraphics[width=\linewidth]{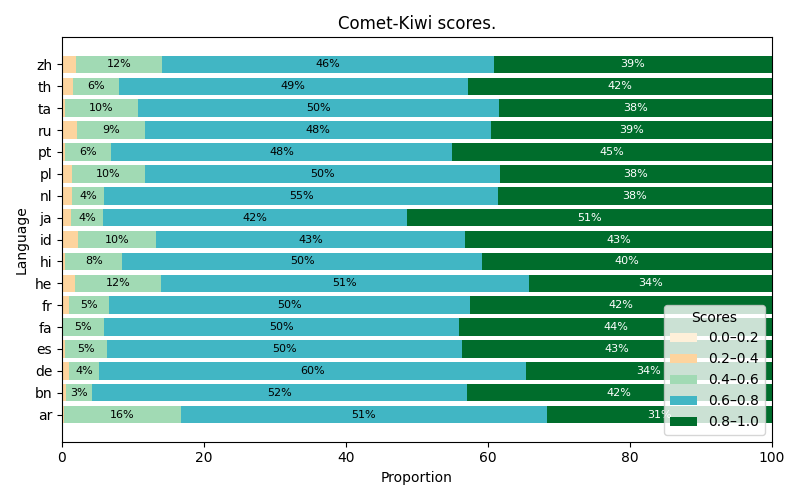}
        \caption{Comet-Kiwi Scores on the translated Answer.}
    \end{subfigure}
    \caption{Translation Quality based on Comet-Kiwi scores.}
    \label{fig:translation_scores}
\end{figure*}

To evaluate the translation qualities, we compute the Comet-Kiwi\cite{rei-etal-2022-cometkiwi} scores for the questions and answers. We group the scores into brackets and visualize them in Figure~\ref{fig:translation_scores}. The majority of the questions get scores $> 0.8$. For the translated answers, the scores are more in the 0.6 - 0.8 bin, probably due to the long sequence length, which makes it harder to determine whether two sequences are translations of each other.

\section{Language Representations}
\begin{figure}[h!]
    \centering
     \begin{subfigure}[b]{0.45\textwidth}
         \centering
         \includegraphics[width=\textwidth]{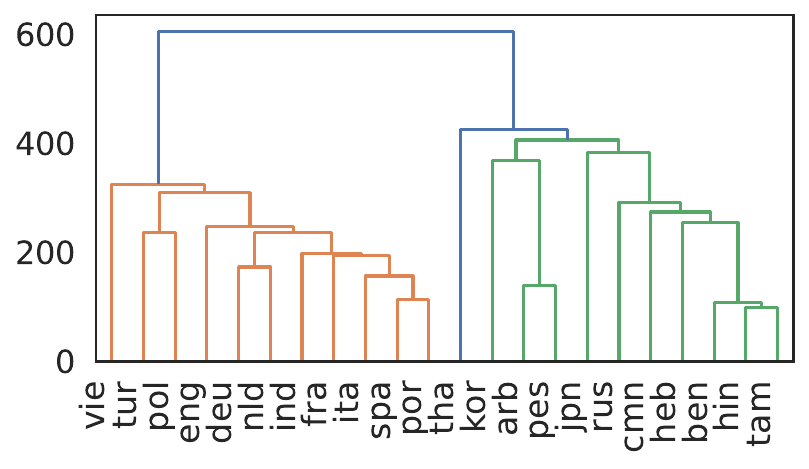}
         \caption{Qwen 2.5}
     \end{subfigure}
     \begin{subfigure}[b]{0.45\textwidth}
         \centering
         \includegraphics[width=\textwidth]{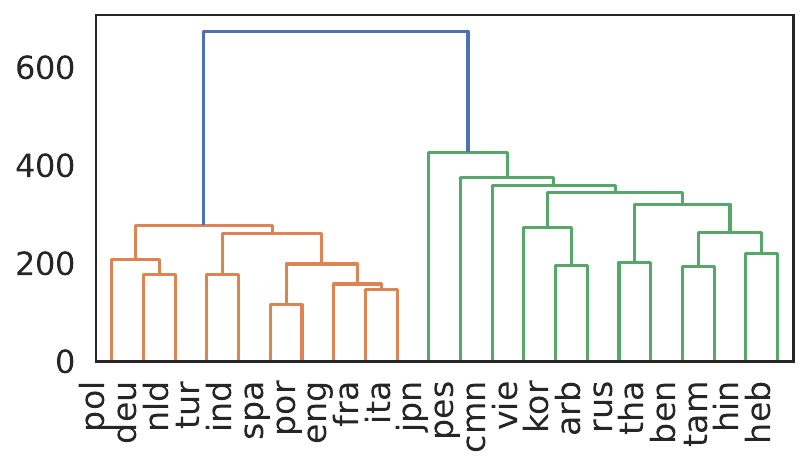}
         \caption{Gemma 2}
     \end{subfigure}
    \caption{Visualization of the language representations of the last layer of Qwen 2.5 and Gemma 2 using Agglomerative Hierarchical Clustering.}
    \label{fig:representation_pca_appendix}
    \vspace{-1mm}
\end{figure}

\begin{table*}[t]
    \small
    \centering
    \begin{tabular}{lllrrrr}
        \toprule
        Lang & Family & Script & Src & Tgt & Mono & Total \\
        \midrule
        ar & Afro-Asiatic & Arabic & 256 & 230 & 51 & 537 \\
        bn & Indo-European & Bengali & 230 & 263 & 65 & 558 \\
        de & Indo-European & Latin & 163 & 244 & 45 & 452 \\
        en & Indo-European & Latin & 207 & 158 & 30 & 395 \\
        es & Indo-European & Latin & 208 & 177 & 29 & 414 \\
        fa & Indo-European & Perso-Arabic & 235 & 221 & 31 & 487 \\
        fr & Indo-European & Latin & 213 & 255 & 63 & 531 \\
        he & Afro-Asiatic & Hebrew & 172 & 119 & 39 & 330 \\
        hi & Indo-European & Devanagari & 206 & 186 & 25 & 417 \\
        id & Austronesian & Latin & 182 & 222 & 35 & 439 \\
        ja & Austronesian & \tiny Han, Hiragana, Katakana & 308 & 242 & 67 & 617 \\
        nl & Indo-European & Devanagari & 145 & 178 & 40 & 363 \\
        pl & Indo-European & Latin & 193 & 157 & 49 & 399 \\
        pt & Indo-European & Latin & 156 & 221 & 39 & 416 \\
        ru & Indo-European & Cyrillic &182 & 153 & 34 & 369 \\
        ta & Dravidian & Tamil & 178 & 195 & 39 & 412 \\
        th & Kra-Dai & Thai & 250 & 213 & 37 & 500 \\
        zh &Sino-Tibetan & Han & 153 & 203 & 45 & 401 \\
        \midrule
        Total &  & & 3637 & 3637 & 763 & 4400 \\
        \bottomrule
    \end{tabular}
    \caption{The number of samples that have each language as a source or target language in for the crosslingual samples and the number of monolingual samples.}
    \label{tab:dataset_stats}
\end{table*}

\section{Position-Specific Steering}
\label{app:positions}

We explore including positions in the computation of the steering vectors. Instead of averaging across positions, the average hidden state is computed per position. This increases the size of the language vector by $ k$ times, where $k$ is the number of positions we consider. However, results in Table \ref{tab:positional} show that there is no advantage to having this information. In fact, it tends to harm performance.

\begin{table}[h!]
    \centering
    \small
    \setlength{\tabcolsep}{8pt}
    \begin{tabular}{llrr}
        \toprule
         Model & Method & Cross  & Mono \\
         \midrule
         \multirow{3}{*}{LLaMA} & + Ours & 97.9 & 99.1 \\
         & + Ours+Pos & 96.7 & 98.6 \\
         & \cellcolor{lightgray}$\Delta$ & \cellcolor{lightgray}-1.2 & \cellcolor{lightgray}-0.5 \\
         \midrule
         \multirow{3}{*}{Qwen} & + Ours & 95.2 & 97.7 \\
         & + Ours+Pos & 95.8 & 93.3 \\
          &  \cellcolor{lightgray}$\Delta$ &  \cellcolor{lightgray}+0.6 &  \cellcolor{lightgray}-4.4 \\
         \midrule
         \multirow{3}{*}{Gemma} & + Ours & 94.5 & 87.8  \\
         & + Ours+Pos & 89.9 & 89.5 \\
         &  \cellcolor{lightgray}$\Delta$ &  \cellcolor{lightgray}-4.6 &  \cellcolor{lightgray}+1.7\\
         \bottomrule
    \end{tabular}
    \caption{LPR on the LCB with language representations for each position.}
    \label{tab:positional}
\end{table}

\section{Detailed Results}
In the main part of the paper we cover monolingual scenarios and crosslingual summarization only aggregated across all languages. Table \ref{tab:lcb_mono} and Table \ref{tab:multiq_mono} show the monolingual performance for the individual languages. Across both datasets, the LPR consistently surpasses $90\%$ for most evaluations, indicating near-optimal performance.

The crosslingual summarization scores on CrossSum are shown in Table \ref{tab:cross_sum}. The \crosslc LPR is shown in Table \ref{tab:cross_sum_lc}. They illustrate that \name and \nametrain can improve \crosslc while maintaining or improving the summarization performance.
\begin{table}[!h]
\small
    \centering
    \begin{tabular}{lrrrrr}
        \toprule
        Model & es & fr &  tr & sw & AVG \\
        \midrule
        Llama & 100.0 & 0.0 & 96.4 & 3.0 & $49.8_{\pm0.004}$ \\
        + LSI & 100.0 & 10.2 & 71.1 & 3.0 & $46.1_{\pm0.003}$ \\
        + \vonename & 100.0 & 100.0 & 98.8 & 85.2 & $96.0_{\pm0.002}$ \\
        % + \vonename & 100.0 & 100.0 & 100.0 & 94.8 & 98.7 \\
        + \mlpname & 100.0 & 100.0 & 100.0 & 95.6 & $98.9_{\pm0.003}$ \\
        \midrule
        Qwen & 98.3 & 91.8 & 95.1 & 9.6 & $73.7_{\pm0.007}$ \\
        + LSI & 97.4 & 75.5 & 92.7 & 20.0 & $71.4_{\pm0.009}$ \\
        + \vonename & 98.8 & 100.0 & 98.8 & 94.8 & $98.1_{\pm0.006}$ \\
        
        % + \vonename & 98.4 & 100.0 & 97.6 & 54.8 & 87.7 \\
        + \mlpname & 100.0 & 100.0 & 100.0 & 91.9 & $98.0_{\pm0.002}$ \\
        \midrule
         Gemma 2 &  84.2 & 89.8 & 30.1 & 0.0 & $51.0_{\pm0.009}$ \\
        + LSI & 98.7 & 99.0 & 4.8 & 0.0 & $50.6_{\pm0.003}$ \\
        + \vonename & 100.0 & 100.0 & 98.8 & 41.0 & $85.0_{\pm0.005}$ \\
        
        % + \vonename & 100.0 & 100.0 & 95.2 & 0.0 & 73.8 \\
        + \mlpname & 100.0 & 100.0 & 100.0 & 52.6 & $88.1_{\pm0.004}$ \\
        \bottomrule
    \end{tabular}
    \caption{\crosslc results on CrossSum for the language pairs: $en \rightarrow es$, $es \rightarrow fr$, $fr \rightarrow tr$, $tr \rightarrow sw$, denoted by the target language, and the standard error for the average across languages.}
    \label{tab:cross_sum_lc}
\end{table}

\begin{table*}[!ht]\centering
\resizebox{\linewidth}{!}{
\begin{tabular}{lrrrrrrrrrrrrrrrrrrrrrr}\toprule
 &\multicolumn{4}{c}{$en \rightarrow es$} &  \multicolumn{4}{c}{$es \rightarrow fr$} & 	\multicolumn{4}{c}{$ fr \rightarrow tr$}  & \multicolumn{4}{c}{$tr\rightarrow sw$} & \multicolumn{4}{c}{AVG} \\
Model & R-1 & R-2 & R-L & Bert & R-1 & R-2 & R-L & Bert & R-1 & R-2 & R-L & Bert & R-1 & R-2 & R-L & Bert & R-1 & R-2 & R-L & Bert  \\
\midrule
\rowcolor{gray!10}
LLama 3.1 & 0.29 & 0.08 & 0.20 & 0.72 & 0.17 & 0.03 & 0.13 & 0.69 & 0.20 & 0.07 & 0.14 & 0.53 & 0.04 & 0.01 & 0.04 & 0.60 & $0.18_{\pm0.002}$ & $0.05_{\pm0.001}$ & $0.13_{\pm0.002}$ & $0.63_{\pm0.002}$
\\
%\midrule
+ LSI &  0.28 &  0.07 &  0.19 &  0.71 &  0.15 &  0.04 &  0.11 &  0.67 &  0.16 &  0.04 &  0.11 &  0.45 &  0.04 &  0.01 &  0.04 &  0.60 &  $0.16_{\pm0.003}$ & $0.04_{\pm0.001}$ & $0.11_{\pm0.002}$ & $0.61_{\pm0.002}$\\
+ \vonename &  0.29 &  0.08 &  0.20 &  0.71 &  0.23 &  0.07 &  0.16 &  0.69 &  0.19 &  0.05 &  0.15 &  0.52 &  0.08 &  0.01 &  0.07 &  0.61 & $0.20_{\pm0.003}$ & $0.05_{\pm0.002}$ & $0.14_{\pm0.002}$ & $0.63_{\pm0.002}$ \\
% + v1 &  0.30 &  0.09 &  0.20 &  0.72 &  0.21 &  0.06 &  0.14 &  0.69 &  0.20 &  0.06 &  0.15 &  0.53 &  0.17 &  0.03 &  0.13 &  0.69 \\
% + v2 & \\
+ \mlpname &   0.30 &  0.09 &  0.21 &  0.72 &  0.22 &  0.07 &  0.15 &  0.69 &  0.20 &  0.06 &  0.15 &  0.54 &  0.14 &  0.02 &  0.10 &  0.66 & $0.22_{\pm0.003}$ & $0.06_{\pm0.002}$ & $0.15_{\pm0.002}$ & $0.65_{\pm0.002}$\\ 
\midrule
\rowcolor{gray!10}
Qwen 2.5 &   0.29 &  0.07 &  0.19 &  0.71 &  0.24 &  0.07 &  0.16 &  0.70 &  0.19 &  0.05 &  0.14 &  0.49 &  0.05 &  0.01 &  0.04 &  0.61 & $0.19_{\pm0.003}$ & $0.05_{\pm0.002}$ & $0.13_{\pm0.002}$ & $0.63_{\pm0.002}$ \\
%\midrule
+ LSI & 0.28 &  0.07 &  0.18 &  0.71 &  0.21 &  0.04 &  0.14 &  0.69 &  0.19 &  0.04 &  0.13 &  0.48 &  0.06 &  0.01 &  0.05 &  0.61 & $0.18_{\pm0.002}$ & $0.04_{\pm0.001}$ & $0.13_{\pm0.002}$ & $0.62_{\pm0.001}$\\
+ \vonename &  0.29 &  0.08 &  0.19 &  0.71 &  0.25 &  0.07 &  0.17 &  0.71 &  0.18 &  0.05 &  0.13 &  0.49 &  0.13 &  0.01 &  0.10 &  0.67  & $0.21_{\pm0.003}$ & $0.05_{\pm0.002}$ & $0.15_{\pm0.002}$ & $0.65_{\pm0.002}$\\
% + v1 &  0.28 &  0.07 &  0.18 &  0.71 &  0.25 &  0.07 &  0.17 &  0.70 &  0.19 &  0.05 &  0.13 &  0.50 &  0.09 &  0.01 &  0.08 &  0.64^\\
% + v2 & \\
+ \mlpname &  0.30 &  0.09 &  0.20 &  0.72 &  0.25 &  0.08 &  0.18 &  0.72 &  0.20 &  0.05 &  0.15 &  0.52 &  0.11 &  0.01 &  0.08 &  0.64 & $0.21_{\pm0.003}$ & $0.06_{\pm0.002}$ & $0.15_{\pm0.002}$ & $0.65_{\pm0.002}$\\
\midrule
\rowcolor{gray!10}
Gemma 2 & 0.26 & 0.07 & 0.18 & 0.71 & 0.23 & 0.06 & 0.17 & 0.71 & 0.12 & 0.03 & 0.10 & 0.41 & 0.05 & 0.01 & 0.05 & 0.61 &  $0.17_{\pm0.003}$ & $0.04_{\pm0.002}$ & $0.12_{\pm0.002}$ & $0.61_{\pm0.002}$
 \\
+ LSI &  0.27 &  0.07 &  0.19 &  0.71 &  0.19 &  0.04 &  0.13 &  0.69 &  0.07 &  0.01 &  0.06 &  0.35 &  0.04 &  0.01 &  0.04 &  0.61 &  $0.15_{\pm0.002}$ & $0.03_{\pm0.001}$ & $0.10_{\pm0.002}$ & $0.59_{\pm0.002}$\\
+ \vonename &  0.29 &  0.08 &  0.20 &  0.71 &  0.25 &  0.07 &  0.18 &  0.71 &  0.19 &  0.05 &  0.14 &  0.52 &  0.05 &  0.01 &  0.04 &  0.57 &  $0.20_{\pm0.003}$ & $0.05_{\pm0.002}$ & $0.14_{\pm0.002}$ & $0.63_{\pm0.002}$ \\ 
% + v1 & 0.29 & 0.08 & 0.20 & 0.72 & 0.25 & 0.08 & 0.18 & 0.71 & 0.20 & 0.06 & 0.15 & 0.52 & 0.05 & 0.01 & 0.05 & 0.61 \\
%+ v2 & \\
+ \mlpname &  0.29 &  0.08 &  0.19 &  0.71 &  0.26 &  0.07 &  0.18 &  0.71 &  0.16 &  0.05 &  0.12 &  0.50 &  0.05 &  0.01 &  0.04 &  0.53 & $0.19_{\pm0.003}$ & $0.05_{\pm0.002}$ & $0.13_{\pm0.002}$ & $0.61_{\pm0.002}$\\
\bottomrule
\end{tabular}}
\caption{The crosslingual summarization performance in Rouge-1, Rouge-2, Rouge-L, and Bert Score on CrossSum with the standard error for the average across languages.}
\label{tab:cross_sum}
\end{table*}

\begin{table*}[!ht]\centering
\resizebox{\linewidth}{!}{
\begin{tabular}{lrrrrrrrrrrrrrrrrr}\toprule
Model  & ar & de & en & es & fr & hi & id & it  & ja & ko & pt & ru & tr  & vi & zh &  avg \\
\midrule
\rowcolor{gray!10}
LLama 3.1 & 99.7 & 100.0 & 98.5 & 99.0 & 100.0 & 100.0 & 93.0 & 100.0 & 99.0 & 100.0 & 95.5 & 100.0 & 99.0 & 100.0 & 96.5 & 98.7 \\
%\midrule
+ LSI & 100.0 & 100.0 & 99.0 & 99.7 & 99.7 & 100.0 & 93.0 & 100.0 & 99.0 & 100.0 & 99.0 & 100.0 & 99.0 & 99.0 & 98.0 & 99.0 \\
+ \neuron & 100.0 & 100.0 & 100.0 & 99.7 & 100.0 & 99.0 & 95.0 & 100.0 & 99.0 & 100.0 & 97.5 & 100.0 & 100.0 & 100.0 & 98.0 & 99.2 \\
+ \vonename & 99.3 & 100.0 & 99.5 & 99.7 & 99.3 & 99.0 & 97.0 & 99.0 & 100.0 & 99.0 & 98.5 & 100.0 & 97.0 & 100.0 & 99.0 & \textbf{99.1} \\
% + \vonename & 99.3 & 99.0 & 98.5 & 98.3 & 100.0 & 100.0 & 91.2 & 99.0 & 100.0 & 100.0 & 96.4 & 100.0 & 88.2 & 100.0 & 97.5 & 97.8 \\
% + v2 & 97.6 & 100.0 & 99.0 & 99.0 & 100.0 & 100.0 & 90.0 & 100.0 & 97.0 & 98.0 & 98.5 & 99.0 & 99.0 & 100.0 & 99.5 & 98.4 \\
+ \mlpname & 99.7 & 100.0 & 99.5 & 98.7 & 100.0 & 100.0 & 94.0 & \cellcolor{cellground}99.0 & 100.0 & \cellcolor{cellground}100.0 & 96.5 & 100.0 & 100.0 & \cellcolor{cellground}99.0 & 99.5 & \textbf{99.1}\\ 
\midrule
\rowcolor{gray!10}
Qwen 2.5 & 98.0 & 99.0 & 100.0 & 98.7 & 98.7 & 100.0 & 98.0 & 100.0 & 92.0 & 97.9 & 96.0 & 100.0 & 98.0 & 100.0 & 98.0 & 98.3 \\
%\midrule
+ LSI & 98.3 & 98.0 & 100.0 & 99.3 & 99.0 & 100.0 & 93.0 & 100.0 & 96.0 & 96.8 & 95.5 & 99.0 & 97.0 & 100.0 & 97.5 & 98.0\\
+ \neuron & 96.7 & 97.0 & 100.0 & 98.7 & 99.0 & 91.0 & 94.0 & 100.0 & 89.0 & 100.0 & 94.5 & 94.8 & 90.9 & 99.0 & 98.5 & 96.2 \\
+ \vonename & 98.7 & 99.0 & 100.0 & 99.3 & 98.3 & 99.0 & 96.0 & 100.0 & 91.0 & 99.0 & 95.5 & 95.0 & 97.0 & 100.0 & 97.5 & 97.7 \\
% + \vonename & 99.0 & 99.0 & 99.5 & 97.7 & 97.0 & 100.0 & 90.0 & 99.0 & 95.0 & 100.0 & 96.5 & 97.9 & 97.0 & 100.0 & 97.0 & 97.6\\
% + v2 & 99.0 & 100.0 & 100.0 & 98.3 & 97.6 & 99.0 & 91.0 & 100.0 & 96.0 & 99.0 & 96.0 & 99.0 & 100.0 & 99.0 & 97.5 & 98.1 \\
+ \mlpname & 99.3 & 98.0 & 100.0 & 99.7 & 99.7 & 98.0 & 96.0 & \cellcolor{cellground}98.0 & 96.0 & \cellcolor{cellground}100.0 & 96.5 & 100.0 & \cellcolor{cellground}99.0 & \cellcolor{cellground}99.0 & 99.0 & \textbf{98.5} \\
\midrule
\rowcolor{gray!10}
Gemma 2 & 83.2 & 98.0 & 98.5 & 95.7 & 94.3 & 67.0 & 71.0 & 94.0 & 99.0 & 99.0 & 97.0 & 93.0 & 99.0 & 80.0 & 57.5 & 88.4 \\
%\midrule
+ LSI & 92.7 & 96.0 & 98.5 & 97.7 & 97.3 & 87.0 & 68.0 & 97.0 & 91.0 & 97.0 & 92.0 & 95.0 & 94.0 & 84.0 & 66.0 & 90.2\\
+ \neuron & 85.2 & 93.9 & 99.5 & 97.7 & 93.3 & 81.0 & 58.0 & 97.0 & 96.0 & 100.0 & 96.5 & 93.0 & 99.0 & 85.0 & 70.0 & 89.7 \\
+ \vonename & 82.9 & 97.0 & 98.5 & 95.3 & 93.0 & 66.0 & 70.0 & 95.0 & 95.0 & 100.0 & 96.0 & 93.0 & 100.0 & 78.0 & 56.5 & 87.8\\
% + \vonename & 76.2 & 96.0 & 98.5 & 94.3 & 92.0 & 63.0 & 68.0 & 95.0 & 96.0 & 99.0 & 96.0 & 91.0 & 99.0 & 76.0 & 58.0 & 86.5 \\
% + v2 & 83.3 & 97.0 & 98.5 & 95.3 & 94.3 & 65.0 & 70.0 & 95.0 & 97.0 & 100.0 & 96.5 & 93.0 & 99.0 & 81.0 & 56.5 & 88.1 \\
+ \mlpname & 98.0 & 100.0 & 100.0 & 99.0 & 99.3 & 100.0 & 90.0 & \cellcolor{cellground}100.0 & 98.0 & \cellcolor{cellground}100.0 & 96.5 & 99.0 & \cellcolor{cellground}100.0 & \cellcolor{cellground}99.0 & 93.0 & \textbf{98.1} \\
\bottomrule
\end{tabular}}
\caption{\monolc results on the LCB. For our learned steering function (\mlpname), languages not seen during training are highlighted in (darker) grey.}
\label{tab:lcb_mono}
\end{table*}

\begin{table*}[!ht]
\small
\centering
%\resizebox{\linewidth}{!}{
\begin{tabular}{lrrrrrrrrrrrr}\toprule
 &\multicolumn{2}{c}{de} &  \multicolumn{2}{c}{en} & \multicolumn{2}{c}{fr} & \multicolumn{2}{c}{tr} & 	\multicolumn{2}{c}{zh} &  \multicolumn{2}{c}{AVG} \\
Model & LPR & Acc & LPR & Acc & LPR & Acc & LPR & Acc & LPR & Acc & LPR & Acc  \\
\midrule
\rowcolor{gray!10}
LLama 3.1 & 96.4 & 65.5 & 94.3 & 67.0 & 95.4 & 65.0 & 98.9 & 57.0 & 87.5 & 67.5 & 94.5 & \textbf{64.4} \\
%\midrule
LSI & 97.4 & 55.0 & 93.8 & 59.5 & 98.0 & 53.5 & 99.4 & 42.0 & 90.0 & 54.0 & 95.7 & 52.8 \\
+ \vonename & 94.4 & 64.0 & 93.9 & 60.5 & 93.9 & 61.0 & 96.1 & 52.5 & 90.9 & 68.5 & 93.8 & 61.3 \\
+ \mlpname & 97.8 & 68.0 & 96.3 & 70.0 & 97.9 & 67.0 & \cellcolor{cellground}95.8 & \cellcolor{cellground}47.0 & 91.0 & 58.5 & \textbf{95.8} & 62.1\\ 
\midrule
\rowcolor{gray!10}
Qwen 2.5 &  90.5 & 67.5 & 92.5 & 59.5 & 90.5 & 58.0 & 89.5 & 35.0 & 90.5 & 89.0 & 90.7 & 61.8 \\
%\midrule
LSI & 93.0 & 43.5 & 93.5 & 59.0 & 93.5 & 52.5 & 92.2 & 22.5 & 91.5 & 77.5 & 92.7 & 51.0 \\
+ \vonename & 90.0 & 60.0 & 95.5 & 68.5 & 92.5 & 67.0 & 91.2 & 38.0 & 91.0 & 80.0 & 92.0 & 62.7 \\
+ \mlpname & 94.3 & 69.0 & 93.4 & 72.5 & 94.2 & 70.5 & \cellcolor{cellground}94.9 & \cellcolor{cellground}40.5 & 89.4 & 80.0 & \textbf{93.2} & \textbf{66.5}\\
\midrule
\rowcolor{gray!10}
Gemma 2 & 91.4 & 29.5 & 93.5 & 30.0 & 92.5 & 42.0 & 96.7 & 42.5 & 84.9 & 48.5 & 91.8 & 38.5
 \\
LSI & 94.5 & 28.0 & 95.5 & 36.5 & 96.5 & 36.5 & 96.8 & 32.5 & 84.5 & 40.5 & \textbf{93.5} & 34.8\\
+ \vonename & 91.9 & 26.0 & 93.5 & 31.5 & 93.0 & 44.0 & 96.7 & 43.0 & 82.9 & 47.0 & 91.6 & 38.3 \\
+ \mlpname & 92.4 & 45.0 & 94.5 & 47.0 & 93.9 & 50.0 & \cellcolor{cellground}92.9 & \cellcolor{cellground}42.0 & 90.8 & 53.0 & 92.9 & \textbf{47.4}\\
\bottomrule
\end{tabular}
\caption{The LPR and Accuracy on the monolingual setup of the MultiQ.}
\label{tab:multiq_mono}
\end{table*}

\begin{table*}[]
    \small
    \centering
    \begin{tabular}{lrrrrrrrrrrrrrr}
        \toprule
         & \multicolumn{7}{c}{Monolingual} & \multicolumn{7}{c}{Crosslingual}\\
         & ar & hi & ja & ko & ru & zh & avg & ar & hi & ja & ko & ru & zh & avg \\
        \cmidrule(lr){2-8}\cmidrule{9-15}
        \rowcolor{gray!10}
        LLAMA & 98.3 & 100.0 & 100.0 & 99.0 & 96.0 & 98.5 & 98.4 & 94.0 & 97.2 & 93.4 & 93.5 & 93.4 & 96.9 & 94.7 \\
        LSI & 99.3 & 100.0 & 100.0 & 98.0 & 97.0 & 96.9 & 98.5 & 91.9 & 99.2 & 94.8 & 97.1 & 96.0 & 97.3 & 96.0 \\
        \vonename & 98.0 & 100.0 & 98.0 & 96.0 & 97.0 & 97.0 & 97.7 & 97.9 & 98.3 & 93.8 & 95.4 & 98.6 & 96.6 & 96.7 \\
        \mlpname & 99.7 & 100.0 & 100.0 & 100.0 & 100.0 & 99.5 & \textbf{99.9}  & 94.7 & 98.3 & 95.9 & 96.5 & 98.6 & 97.5 & \textbf{96.9} \\
        \midrule
        \rowcolor{gray!10}
        Qwen & 100.0 & 99.0 & 98.9 & 100.0 & 98.0 & 99.0 & 99.2  &  94.3 & 96.7 & 89.4 & 95.5 & 93.5 & 96.2 & 94.3 \\
        LSI & 100.0 & 100.0 & 99.0 & 98.9 & 100.0 & 100.0 & \textbf{99.6} &  94.3 & 98.1 & 89.9 & 95.7 & 94.4 & 94.8 & 94.5 \\
        \vonename &  99.7 & 100.0 & 98.9 & 100.0 & 95.7 & 99.5 & 99.0 & 96.2 & 98.3 & 91.7 & 95.3 & 94.8 & 88.0 & 94.0 \\
        \mlpname & 100.0 & 99.0 & 100.0 & 99.0 & 100.0 & 99.5 & \textbf{99.6} & 96.7 & 97.5 & 93.8 & 94.8 & 97.1 & 98.4 & \textbf{96.4} \\
        \midrule
        \rowcolor{gray!10}
        Gemma & 96.7 & 100.0 & 91.9 & 100.0 & 93.6 & 96.7 & 96.5 & 86.2 & 98.8 & 98.7 & 90.9 & 93.5 & 98.6 & \textbf{94.5} \\
        LSI & 96.6 & 98.9 & 100.0 & 97.9 & 96.8 & 96.1 & 97.7 & 69.0 & 91.4 & 88.0 & 86.6 & 88.4 & 97.6 & 86.8 \\
        \vonename & 95.6 & 98.5 & 94.7 & 100.0 & 94.6 & 96.5 & 96.7& 91.0 & 97.2 & 83.6 & 94.4 & 94.2 & 91.8 & 92.0 \\
        \mlpname & 98.6 & 100.0 & 100.0 & 100.0 & 96.0 & 100.0 & \textbf{99.1} & 86.2 & 96.7 & 92.9 & 95.9 & 97.6 & 92.6 & 93.6 \\ 
        \bottomrule
    \end{tabular}
    \caption{WPR on the monolingual and crosslingual portion of the LCB. }
    \label{tab:lcb_wpr}
\end{table*}

\end{document}